\pdfoutput=1
\documentclass{article}
\usepackage[nonatbib, final]{neurips_2020}

\usepackage[T1]{fontenc}
\usepackage[hyperref]{xcolor}
\usepackage{microtype}

\usepackage{math}
\usepackage{xkcd_color}

\usepackage[
    colorlinks=true,
    linkcolor={xkcd red},
    citecolor={xkcd blue},
    urlcolor={xkcd magenta},
    backref=section,
]{hyperref}

\usepackage{natbib}
\usepackage{booktabs}
\usepackage{multirow}
\usepackage[flushleft]{threeparttable}

\usepackage{amsthm}
\newtheorem{theorem}{Theorem}
\newtheorem{assumption}[theorem]{Assumption}
\newtheorem{proposition}[theorem]{Proposition}
\theoremstyle{definition}
\newtheorem{definition}[theorem]{Definition}

\usepackage{enumitem} % alphabet enumerate
\usepackage{siunitx} % scientific notation

\title{Learning from Aggregate Observations}

\author{
  Yivan Zhang\\
  The University of Tokyo / RIKEN \\
  \texttt{yivanzhang@ms.k.u-tokyo.ac.jp} \\
  \And
   Nontawat Charoenphakdee \\
  The University of Tokyo / RIKEN \\
  \texttt{nontawat@ms.k.u-tokyo.ac.jp} \\
   \AND
   Zhenguo Wu \\
   The University of Tokyo \\
   \texttt{zhenguo@ms.k.u-tokyo.ac.jp} \\
  \And
   Masashi Sugiyama \\
   RIKEN / The University of Tokyo \\
   \texttt{sugi@k.u-tokyo.ac.jp} \\
}

\begin{document}

\maketitle

\begin{abstract}
We study the problem of \emph{learning from aggregate observations} where supervision signals are given to \emph{sets} of instances instead of individual instances, while the goal is still to predict labels of unseen individuals.
A well-known example is multiple instance learning (MIL). 
In this paper, we extend MIL beyond binary classification to other problems such as multiclass classification and regression.
We present a general probabilistic framework that accommodates a variety of aggregate observations, e.g., pairwise similarity/triplet comparison for classification and mean/difference/rank observation for regression.
Simple maximum likelihood solutions can be applied to various differentiable models such as deep neural networks and gradient boosting machines.
Moreover, we develop the concept of \emph{consistency up to an equivalence relation} to characterize our estimator and show that it has nice convergence properties under mild assumptions.
Experiments on three problem settings --- \emph{classification via triplet comparison} and \emph{regression via mean/rank observation} indicate the effectiveness of the proposed method.

\end{abstract}

\section{Introduction}
Modern machine learning techniques usually require a large number of high-quality labels for \emph{individual} instances \citep{jordan2015machine}.
However, in the real world, it could be prohibited to obtain individual information while we can still obtain some forms of supervision for \emph{sets} of instances.
We give three following motivating use cases.
The first scenario is when individual information cannot be released to the public due to \emph{privacy concerns} \citep{horvitz2015data}.
One way to avoid releasing individual information is to only disclose some summary statistics of small groups to the public so that individual information can be protected.
The second scenario is when individual annotations are \emph{intrinsically unavailable} but group annotations are provided, which arises in problems such as drug activity prediction problem \citep{dietterich1997solving}.
The third scenario is when the labeling cost for an individual instance is \emph{expensive} \citep{zhou2017brief}.

In classification, one of the most well-known examples of learning from aggregate observations is \emph{multiple instance learning} (MIL) for binary classification \citep{zhou2004multi}, where training instances are arranged in sets and the aggregate information is whether the positive instances exist in a set.
MIL has found applications in many areas \citep{yang2005review, carbonneau2018multiple}, including aforementioned drug activity prediction \citep{dietterich1997solving}.
\emph{Learning from label proportions} (LLP) \citep{kuck2012learning, quadrianto2009estimating, felix2013psvm, patrini2014almost} is another well-studied example where the proportion of positive instances in each set is observed.
However, most earlier studies as well as recent work in MIL and LLP only focus on binary classification and the type of aggregation is limited.
Recently, \emph{classification from comparison} has gained more attention. 
\citet{bao2018classification} considered \emph{learning from pairwise similarities}, where one can obtain a binary value indicating whether two instances belong to the same category or not.
\citet{cui2020classification} considered \emph{learning from triplets comparison}. 
Nevertheless, both works~\citep{bao2018classification, cui2020classification} are only applicable to binary classification. For multiclass classification from aggregate observations, we are only aware of a recent work by \citet{hsu2019multiclass}, where pairwise similarities are used to train a multiclass classifier based on maximum likelihood estimation, although its theoretical understanding is limited.

In regression, learning from aggregate observations has been assessed to only a limited extent.
One example is \emph{multiple instance regression} \citep{ray2001multiple, amar2001multiple}, where it is assumed that there is only one instance in each set, called the ``primary instance'', that is responsible for the real-valued label.
Another example is \emph{learning from spatially aggregated data}, where only spatially summed or averaged values can be observed. 
Gaussian processes (GP) were proposed to handle this problem \citep{law2018variational,tanaka2019refining,tanaka2019spatially}.
Although uncertainty information can be nicely obtained with a GP-based method, it may not be straightforward to scale it for high-dimensional data. 

In general, the concept of drawing conclusions or making predictions about individual-level behavior based on aggregate-level data was also of great interest in other fields, such as ecological inference \citep{schuessler1999ecological, king2004ecological, sheldon2011collective, flaxman2015supported} and
preference learning \citep{thurstone1927law, bradley1952rank, luce1959individual, plackett1975analysis}.

The goal of this paper is to provide a versatile framework that can be applied for many types of aggregate observations, both in multiclass classification and regression.
We propose a simple yet effective solution to this class of problems based on maximum likelihood estimation, where \citet{hsu2019multiclass}'s method is one of our special cases. 
Note that the maximum likelihood method has been explored for weak supervisions such as classification from noisy labels and coarse-grained labels \citep{patrini2017making, zhang2019learning}, but only~\citet{hsu2019multiclass} considered aggregate observations.

Our contributions can be summarized as follows. 
Firstly, we expand the usage of maximum likelihood to aggregate observations, e.g., classification from triplet comparison and regression from mean/rank observation.
Next, we provide a theoretical foundation of this problem by introducing the concept of \emph{consistency up to an equivalence relation} to analyze the characteristics of our estimator.
We demonstrate that this concept is highly useful by analyzing the behavior of our estimator and obtain insightful results, e.g., regression from mean observation is consistent but regression from rank observation is only consistent \emph{up to an additive constant}.
Finally, experimental results are also provided to validate the effectiveness of our method.

\section{Problem setting}
\label{sec:problem}
\begin{figure}[t]
\centering
\begin{tikzpicture}
\platenotation
\node (X) [observable] {$X$};
\node (Z) [unobservable, right = of X] {$Z$};
\node (Y) [observable, right = of Z] {$Y$};
\path (X) edge [dependency] (Z);
\path (Z) edge [dependency] (Y);
\node (W) [unobservable, above = 8mm of Z] {$W$};
\path (W) edge [dependency] (Z);
\node [plate, fit=(X)(Z)(Y), inner sep = 5mm,
       label={[anchor=south east] south east: $N$}] {};
\node [plate, fit=(X)(Z), inner sep = 3mm,
       label={[anchor=south east] south east: $K$}] {};
\end{tikzpicture}
\caption{\textbf{Graphical representation of the data generating process.} 
$X$ represents the \emph{feature vector} and $Z$ represents the unobservable \emph{true target} for each $X$. 
An \emph{aggregate observation} $Y$ can be observed from a set of $Z$, $Z_{1:K}$, via an \emph{aggregate function} $T: \sZ^K \mapsto \sY$, where $K \geq 2$ is the cardinality of the set.
$Z$ follows a parametric distribution $p(Z | \theta)$ parameterized by $\theta = f(X; W)$, which is the output of a deterministic function $f$, parameterized by $W$.
The goal is to estimate $W$ from $(X_{1:K}, Y)$-pairs to predict the true target $Z$ from a single feature vector $X$.
}
\label{fig:xzy}
\vspace{-3mm}
\end{figure}
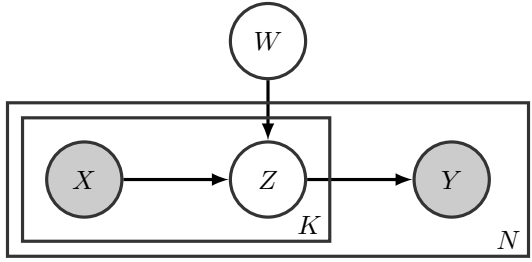

\begin{table*}[t]
\centering
\vspace{-2em}
\caption{\textbf{Examples of Learning from Aggregate Observations}}
\label{tab:examples}
\resizebox{\textwidth}{!}{%
\begin{tabular}{llll}
\toprule
Task 
& Learning from \dots 
& $K$
& Aggregate Observation 
$ Y = T(Z_{1:K}) $ 
\\
\midrule
\multirow{5}{*}{
\begin{tabular}[c]{@{}l@{}}
  Classification\\ 
  $ Z \in \{1, \dots, C\} $
\end{tabular}} 
& similarity/dissimilarity 
& $K = 2$
& if $Z_1$ and $Z_2$ are the same or not
\\
\cmidrule{2-4}
& triplet comparison 
& $K = 3$
& 
\begin{tabular}[c]{@{}l@{}}
  if $ d(Z_1, Z_2) $ is smaller than $ d(Z_1, Z_3) $,\\ 
  where $ d(\cdot, \cdot) $ is a similarity measure between classes
\end{tabular}
\\
\cmidrule{2-4}
& multiple instance 
& $K \geq 2$
& if $Z_{1:K}$ contains positive instances ($C = 2$)
\\
%%%%%%%%%%%%%%%%%%%%%%%%%%%%%%%%%%%%%%
\midrule
\multirow{5}{*}{
\begin{tabular}[c]{@{}l@{}}
  Regression\\
  $ Z \in \R $
\end{tabular}} 
& mean/sum 
& $K \geq 2$
& the arithmetic mean or the sum of $Z_{1:K}$
\\ 
\cmidrule{2-4}
& difference/rank 
& $K = 2$
& the difference $ Z_1 - Z_2 $, or the relative order $ Z_1 > Z_2 $
\\
\cmidrule{2-4}
& min/max 
& $K \geq 2$
& the smallest/largest value in $Z_{1:K}$
\\
\cmidrule{2-4}
& uncoupled data 
& $K \geq 2$
& randomly permuted $Z_{1:K}$
\\
\bottomrule
\end{tabular}
} % \resizebox
\vspace{-4mm}
\end{table*}

%%%%%%%%%%%%%%%%%%%%%%%%%%%%%%%%%%%%%%%%%%%%%%%%%%

In this section, we clarify the notation and assumptions and introduce a probabilistic model for aggregate observations.

%%%%%%%%%%%%%%%%%%%%%%%%%%%%%%%%%%%%%%%%%%%%%%%%%%

\newpage
\subsection[Notation]{Notation\footnotemark}
\footnotetext{%
In this work, we denote uppercase letters $X, Y, Z$ as random variables, lowercase letters $x, y, z$ as instances of random variables, and calligraphic letters $\sX, \sY, \sZ$ as their support spaces.
The subscript such as $Z_{1:K}$ is an abbreviation for the set $\{Z_1, Z_2, \dots, Z_K\}$.
The superscript such as $y^{(i)}$ denotes the $i$-th sample point in a dataset.
With abuse of notation, $p(\cdot)$ denotes a distribution and also its probability mass/density function.
}

Let $ X \in \sX $ be the \emph{feature vector}, and $ Z \in \sZ $ be the unobservable \emph{true target} that we want to predict, where $\sX$ and $\sZ$ are their support spaces, respectively.
The goal is to learn a discriminative model that predicts the true target $Z$ from the feature vector $X$.
We do not have any restrictions on $\sZ$ as long as we can model the conditional probability $p(Z | X)$.
If $\sZ$ is a finite set, e.g., $\{1, \dots, C\}$, the task is $C$-class classification; and if $\sZ$ is $\R$, the task is regression.
However, in the problem setting of learning from aggregate observations, we cannot observe the true target $Z$ directly.
Instead, we assume that we can observe some information about \emph{multiple instances}.

Concretely, the supervision signal $Y \in \sY$, called \emph{aggregate observation}, for the set $Z_{1:K}$, can be obtained from $Z_{1:K}$
via an \emph{aggregate function} $T: \sZ^K \mapsto \sY$, 
i.e., $Y = T(Z_{1:K})$.
We want to use observations of $(X_{1:K}, Y)$ to predict the true target $Z$ based on features $X$.
Figure~\ref{fig:xzy} illustrates the graphical representation of the data generating process.

%%%%%%%%%%%%%%%%%%%%%%%%%%%%%%%%%%%%%%%%%%%%%%%%%%

\subsection{Assumptions}
\label{ssec:assumptions}

Here, we summarize our assumptions used in learning from aggregate observations as follows.

\begin{assumption}[Aggregate observation assumption]
$p(Y | X_{1:K}, Z_{1:K}) = p(Y | Z_{1:K})$.
\end{assumption}

This means that $Y$ and $X_{1:K}$ are conditionally independent given $Z_{1:K}$.
This assumption is common in existing studies \citep{carbonneau2018multiple, hsu2019multiclass, xu2019uncoupled, cui2020classification}.
It can be implied in the data collection process, e.g., when we first collect $(X, Z)$-pairs but only disclose some summary statistics of $Z_{1:K}$ for learning due to privacy concerns.
It also means that we expect the true target $Z$ to carry enough information about $X$, so that we do not need to extract more information from $X_{1:K}$ to predict $Y$.

Further, since we assumed that $Y$ can be determined by 
$Y = T(Z_{1:K})$, 
the conditional probability becomes
$p(Y | Z_{1:K}) = \delta_{T(Z_{1:K})}(Y)$,
where $\delta(\cdot)$ denotes the \emph{Dirac delta function}.

\begin{assumption}[Independent observations assumption]
$p(Z_{1:K} | X_{1:K}) = \prod_{i=1}^K p(Z_i | X_i)$.
\end{assumption}
This means that elements in $Z_{1:K}$ are mutually independent.
It might be violated if we collect the data group by group, so that instances in a group tend to be similar and have high correlation \citep{carbonneau2018multiple}.
In such a case, our formulation and proposed method only serve as a practical approximation and our theoretical implications may not hold.
Note that if the independence condition does not hold, aggregate observations could be more informative in some scenarios.
For example, spatially aggregated data are aggregated within a specific region rather than randomly collected from many different regions~\citep{law2018variational}.
Thus, one may be able to utilize the dependency in $Z_{1:K}$ to obtain a better estimation of $p(Z_{1:K} | X_{1:K})$.

Combining these two assumptions, we can decompose the joint distribution $p(X_{1:K}, Z_{1:K}, Y)$ as
\begin{equation}
\label{eq:decomposition}
  p(X_{1:K}, Z_{1:K}, Y)
=
  p(Y | Z_{1:K}) \prod_{i=1}^K p(Z_i | X_i) p(X_i)
.
\end{equation}

%%%%%%%%%%%%%%%%%%%%%%%%%%%%%%%%%%%%%%%%%%%%%%%%%%

\subsection{Aggregate function}
The aggregate function $ T: \sZ^K \mapsto \sY $ characterizes different problems within our framework.
It induces an \emph{equivalence relation} $\sim$ over the set $\sZ^K$ as
$ 
  z_{1:K}^{(i)} \sim z_{1:K}^{(j)} 
\iff 
  T(z_{1:K}^{(i)}) = T(z_{1:K}^{(j)})
$.
The coarseness of $\sim$ influences how hard the problem is.
Since $\sim$ is usually much coarser than the equality equivalence relation on $\sZ^K$, the model is not able to distinguish some observationally equivalent $Z_{1:K}$.
However, we can analyze this equivalence relation $\sim$ and obtain the characteristics of the solution.
In some problems, we are able to recover the conditional distribution $p(Z | X)$, or at least up to some transformation.
We show this fact theoretically in Section~\ref{sec:consistency} and experimentally in Section~\ref{sec:experiments}.
In Table~\ref{tab:examples}, we provide examples of problems that can be accommodated in our framework of learning from aggregate observations, which are categorized by underlying tasks and their aggregate functions.

\section{Proposed method}
\label{sec:method}
In this section, we describe the proposed method for learning from aggregate observations.

The key idea is to estimate $ p(Y | X_{1:K}) $ based on $ p(Z | X) $ using the aggregate function $T$.
Based on the decomposition of the joint distribution in Equation~\eqref{eq:decomposition}, we can marginalize 
$ 
  p(Y, Z_{1:K} | X_{1:K}) 
= 
  p(Y | Z_{1:K}) 
  \prod_{i=1}^K p(Z_i | X_i) 
$
over $Z_{1:K}$ as
\begin{equation}
\label{eq:pyx}
  p(Y | X_{1:K})
=
  \int_{\sZ^K}
  \delta_{T(z_{1:K})}(Y) 
  \prod_{i=1}^K p(z_i | X_i)
  \D{z_{1:K}}
=
  \E_{\substack{
  Z_i \sim p(Z_i | X_i)\\ 
  i=1, \dots, K}}
  \brackets*{
  \delta_{T(Z_{1:K})}(Y)
  }
,
\end{equation}
where
$\E[\cdot]$ denotes the expectation.

To model the true target $Z$,
 we assume that 
$Z$ conditioned on $X$ follows a certain parametric distribution, parameterized by $ \theta \in \Theta $, i.e., 
$
  p(Z_i | X_i)
=
  p(Z_i | \theta = f(X_i; W))
$.
Here, $ f: \sX \to \Theta $ is a deterministic function parameterized by $ W \in \sW $, which maps the feature vector $X$ to the corresponding parameter $\theta$.
Note that we do not restrict the family of $f$.
It can be a deep neural network~\citep{lecun2015deep} or a gradient boosting machine \citep{friedman2001greedy}, as long as the likelihood $ p(Z | X) $ is differentiable w.r.t.~the model parameter $W$.

Then, we use the likelihood to measure how likely our model with parameter $W$ can generate observed data.
Our learning objective, the \emph{expected log-likelihood}, is defined as
\begin{equation}
  \ell(W)
= 
  \E
  \brackets*{
  \log p(Y | X_{1:K}; W)
  }
.
\end{equation}
Approximating $\ell(W)$ based on an i.i.d.~sample of $ (X_{1:K}, Y) $-pair of size $N$,
$ 
  \braces*{x_{1:K}^{(i)}, y^{(i)}}_{i=1}^N
\iid
  p(X_{1:K}, Y)
$,
the \emph{log-likelihood} is defined as
\begin{equation}
\label{eq:log_likelihood}
  \ell_N(W)
= 
  \frac{1}{N}
  \sum_{i=1}^N
  \log p(y^{(i)} | x_{1:K}^{(i)}; W)
,
\end{equation}
which converges to $\ell(W)$ almost surely as the sample size $N \to \infty$ \citep{van2000asymptotic}.
Then, the \emph{maximum likelihood estimator} (MLE) of $W$ given aggregate observations is defined as
$\widehat{W}_N = \argmax_W \ell_N(W)$.

Although MLE exists for all kinds of aggregate observations in theory, the likelihood can be hard to calculate because of the integral involved in Equation~\eqref{eq:pyx}.
In Section~\ref{sec:realizations}, we give three examples where we can obtain an analytical expression of the log-likelihood so that we do not need to resort to numerical integration or approximation methods to optimize the learning objective.

\section{Consistency of learning from aggregate observations}
\label{sec:consistency}
In this section, we develop theoretical tools and show that although the estimator may not always converge to the true parameter, it can still capture some important information about the true parameter. 
In Section~\ref{sec:realizations}, we show that our theoretical analysis can elucidate the fact that our estimators can have different types of consistency depending on the given aggregate observations. 

Assume that the parameter $W$ is in a metric space $(\sW, d)$ with a metric $d$.
Denote the true parameter by $W_0$.
An estimator $\widehat{W}_N$ based on $N$ sample points is said to be \emph{consistent} 
if $\widehat{W}_N \convergep W_0$ as $N \to \infty$, 
where $\convergep$ means the convergence in probability.

The MLE is consistent under mild conditions \citep{van2000asymptotic}, but the \emph{identifiability} of the model is always necessary.
However, in learning from aggregate observations, it is more common that we can only \emph{partially} identify the model.
i.e., two parameter $W$ and $W'$ might be \emph{observationally equivalent} and yield the same likelihood, which is defined as follows:
\begin{definition}[Equivalence]
\label{def:equivalence}
An \emph{equivalence relation} $\sim$ on $\sW$ induced by the likelihood is defined according to
$W \sim W' \iff \ell(W) = \ell(W')$.
The equivalence class of $W$ is denoted by $[W]$.
\end{definition}

Then, \emph{consistency up to an equivalence relation} is defined as follows:
\begin{definition}[Consistency up to $\sim$]
\label{def:const-up-to}
An estimator $ \widehat{W}_N $ is said to be \emph{consistent up to an equivalence relation $\sim$},
if $ d(\widehat{W}_N, [W_0]) \convergep 0$ 
as $ N \to \infty $,
where $d(W, [W_0]) = \inf_{W_0'\in [W_0]} d(W, W_0')$.
\end{definition}

That is to say, an estimator converges in probability to a set of values that is at least observationally equivalent to the true parameter.
Then, following \citet{wald1949note} and \citet{van2000asymptotic}, we can confirm that the MLE given aggregate observations has the desired behavior under mild assumptions:
\begin{proposition}
\label{prop:condition}
The MLE $\widehat{W}_N$ based on Equation~\eqref{eq:log_likelihood} is consistent up to $\sim$ if the following conditions hold:
\begin{enumerate}
[itemsep=-1mm, topsep=-1mm, label=(\Alph*)]
\item
The parameter space $\sW$ is compact;

\item
$\forall W \in \sW$, the log-likelihood 
$\ell(W | x_{1:K}, y) = \log p(x_{1:K}, y; W)$ 
is upper-semicontinuous for almost all $(x_{1:K},y)$;

\item
$\forall$ sufficiently small ball $U \subset \sW$,
the function
$\ell^U(W | x_{1:K}, y) = \sup_{W\in U}\ell(W|x_{1:K},y)$
is measurable and satisfies 
$\E\brackets*{\ell^U(W | X_{1:K}, Y)} < \infty$;

\item
$\exists W_0' \in [W_0]$
s.t. $\ell_N(\widehat{W}_N) \geq \ell_N(W_0') - \delta_N$, 
where $\delta_N$ is a sequence of random variables such that $\delta_N \convergep 0$.
\end{enumerate}
\end{proposition}

Definitions~\ref{def:equivalence}, \ref{def:const-up-to} and Proposition~\ref{prop:condition} provide a foundation to analyze the behavior of an MLE for learning from aggregate observations.
The key difference between an analysis of the traditional MLE and ours is that due to limited supervision, the concept of \emph{consistency} can be too strict. 
Although an MLE is not consistent for some aggregate observation (see Sections~\ref{ssec:method_triplet} and~\ref{ssec:method_rank}), we can show that an MLE is still useful by using the concept of \emph{consistency up to an equivalence relation}.

\section{Realizations of learning from aggregate observations}
\label{sec:realizations}
In this section, we illustrate three novel realizations of the proposed maximum likelihood approach to learning from aggregate observations where the integral in Equation~\eqref{eq:pyx} can be solved analytically.
Furthermore, we demonstrate how to use the theoretical tools we developed in Section~\ref{sec:consistency} to provide a better understanding of the behavior of the estimator given different kinds of aggregate observations.

Here, we will drop $\theta = f(X; W)$ from expressions to keep the notation uncluttered.
Any parameter of the distribution of $Z$ is either determined from $X$ in this way or just kept fixed.
We defer the proofs of propositions in this section to Appendix~\ref{app:proofs}.

%%%%%%%%%%%%%%%%%%%%%%%%%%%%%%%%%%%%%%%%%%%%%%%%%%

\subsection{Classification via pairwise similarity}
\label{ssec:method_similarity}

In classification, the true target is categorical: 
$Z \in \{1, \dots, C\}$, 
where $C$ is the number of classes.
We model it using a categorical distribution
$Z \sim \Categorical(p_{1:C})$,
with probability parameters
$p_i > 0$, $\sum_{i=1}^C p_i = 1$.
Because $\abs{\sZ^K} = C^K$ is always finite, the integration in Equation~\eqref{eq:pyx} becomes a summation and the log-likelihood is always analytically calculable.

\textbf{Pairwise similarity} is our first example of learning from aggregate observations.
In this problem, we only know if a pair of instances belongs to the same class or not.
Here $ K = 2$.
Concretely,
\begin{equation}
  Y = T_\mathrm{sim}(Z_1, Z_2) = [Z_1 = Z_2]
,
\quad
  p(Y = 1) = \sum_{i=1}^C p(Z_1 = i) p(Z_2 = i)
,
\end{equation}
where $[\cdot]$ denotes the Iverson bracket.\footnote{Iverson bracket $[\cdot]$: for any logic expression $P$, $[P] = 1$ if $P$ is true, otherwise $0$.}
This problem is also considered to be \emph{semi-supervised clustering} in the literature \citep{xing2003distance, basu2004probabilistic, bilenko2004integrating, davis2007information}.
Recently, \citet{hsu2019multiclass} studied this problem thoroughly for several image classification tasks from the perspective of classification, whose method is also based on the maximum likelihood and thus a special case within our framework.

%%%%%%%%%%%%%%%%%%%%%%%%%%%%%%%%%%%%%%%%%%%%%%%%%%

\subsection{Classification via triplet comparison}
\label{ssec:method_triplet}

\textbf{Triplet comparison} is of the form ``$A$ (\emph{anchor}) is more similar to $P$ (\emph{positive}) than $N$ (\emph{negative}).''
Assuming that $d: \sZ \times \sZ \mapsto \R$ is a similarity measure between classes, we can instantiate the aggregation function $T$ and Equation~\eqref{eq:pyx} for $K=3$:
\begin{equation}
\begin{aligned}
  Y = T_\mathrm{tri}(Z_1, Z_2, Z_3) 
= 
  [d(Z_1, Z_2) < d(Z_1, Z_3)]
,
\\
  p(Y = 1) 
= 
  \sum_{
  \substack{d(i, j) < d(i, k)\\ 
  i, j, k \in \{1, \dots, C\}}}
  p(Z_1 = i) p(Z_2 = j) p(Z_3 = k)
.
\end{aligned}
\end{equation}

Based on Equation~\eqref{eq:log_likelihood}, the loss function for triplet comparison is in the form of binary cross entropy:
\begin{equation}
\begin{aligned}
  \ell_N^\mathrm{tri}(W)
=
  &-\frac1N \sum_{i=1}^N
  [y^{(i)} = 1] \log p(y^{(i)} = 1 | x_{1:K}^{(i)}; W)
\\
  &-\frac1N \sum_{i=1}^N
  [y^{(i)} = 0] \log p(y^{(i)} = 0 | x_{1:K}^{(i)}; W)
.
\end{aligned}
\end{equation}

Although triplet comparison data has been used for \emph{metric learning} and \emph{representation learning} \citep{sohn2016improved, schroff2015facenet, mojsilovic2019relative, schultz2004learning, kumar2008semisupervised, tamuz2011adaptively,baghshah2010kernel}, studies on classification based on solely triplet comparison remain limited.
\citet{cui2020classification} recently showed that it is possible to learn a binary classifier from triplet comparison directly.
But so far no result has been given for the multiclass setting, to the best of our knowledge. 
Existing work either requires labeled data~\citep{perrot2019boosting} or needs to actively inquire comparison labels from an oracle~\citep{haghiri2018comparison}. 

Since the parameters of the model given only the pairwise similarity or the triplet comparison are not identifiable, the corresponding MLE cannot be consistent.
But that does not mean the learned model does not capture any information. 
Based on our framework, we prove the following proposition, which suggests that a multiclass classifier is still learnable at most up to a permutation of classes.

\begin{proposition}[Classification via triplet comparison is at most consistent up to a permutation]
\label{prop:triplet}
Let $f(X; W)$ be a $C$-dimensional vector of probability parameters $p_{1:C}$ for classification, then
\begin{equation}
  \braces*{
  W \in \sW: 
  \exists \textnormal{ permutation matrix } P,
  \st
  Pf(X; W) = f(X; W_0) 
  \almosteverywhere
  }
\subseteq
  [W_0] 
.
\end{equation}
\end{proposition}

Proposition~\ref{prop:triplet} states that the estimator can be consistent up to a permutation, but \emph{not always}.
Please refer to Appendix~\ref{app:proofs} for more discussion.
The same result holds for the pairwise similarity case, which was investigated empirically by~\citet{hsu2019multiclass}.
In Section~\ref{ssec:experiment_triplet}, we show that the proposed method works well empirically when the true labels are not too ambiguous.

%%%%%%%%%%%%%%%%%%%%%%%%%%%%%%%%%%%%%%%%%%%%%%%%%%

\subsection{Regression via mean observation}
\label{ssec:method_mean}

In regression problems, the true target is a real value 
$Z \in \R$.
The \emph{mean squared error} (MSE) is the canonical choice for the loss function, which can be derived from the maximum likelihood under an \emph{additive homoscedastic Gaussian noise} model \citep{bishop2006pattern}.
Thus, we also use a Gaussian distribution for the true target,
$Z \sim \Normal(\mu, \sigma^2)$
with mean $\mu \in \R$
and standard deviation $\sigma \in (0, \infty)$.

Gaussian distributions have desired properties, including \emph{stability} and \emph{decomposability}. 
Therefore Gaussian distribution is closed under linear combination given independence (See Appendix~\ref{app:mse}).
We can model several aggregate observations, e.g., the mean observation and the rank observation, and obtain an analytical expression of the log-likelihood, as discussed below.

\textbf{Mean observation} is the arithmetic mean of the set of true targets $Z_{1:K}$.
Under the Gaussian distribution assumption, the mean $Y$ is also a Gaussian random variable:
\begin{equation}
\label{eq:mean}
  Y = T_\mathrm{mean}(Z_{1:K}) = \frac{1}{K}\sum_{i=1}^K Z_i
, \quad
  Y 
\sim 
  \Normal\parens*{
  \frac{1}{K}\sum_{i=1}^K \mu_i, \frac{1}{K^2}\sum_{i=1}^K \sigma_i^2
  }
.
\end{equation}
Assuming homoscedastic noise,
$Z_i = f(X_i; W) + \varepsilon_i$, 
for $i = 1, \dots, K$, 
where $\varepsilon_i \iid \Normal(0, \sigma^2)$,
the loss function realized from Equation~\eqref{eq:log_likelihood} becomes
\begin{equation}
  \ell_N^\mathrm{mean}(W)
=
  \frac1N \sum_{i=1}^N
  \parens*{
  y^{(i)} - \frac{1}{K}\sum_{j=1}^K f(x_j^{(i)}; W)
  }^2
.
\end{equation}
Theoretically, we can obtain the following proposition, which states that our estimator is consistent:
\begin{proposition}[Regression via mean observation is consistent]
\label{prop:mean}
The MLE $\widehat{W}_N$ based on Equation~\eqref{eq:log_likelihood} for mean observations obtained by Equation~\eqref{eq:mean} is consistent, because
\begin{equation}
  [W_0] 
= 
  \{ 
  W \in \sW: 
  f(X; W) = f(X; W_0)
  \almosteverywhere
  \} 
.
\end{equation}
\end{proposition}

In Section~\ref{ssec:experiment_mean_rank}, we validate Proposition~\ref{prop:mean} and demonstrate that it is feasible to learn solely from mean observations using our method and achieve MSE that is comparable to learning from direct observations.
In Appendix~\ref{app:mean}, we provide possible use of the Cauchy distribution for robust regression and the Poisson distribution for count data in addition to the Gaussian distribution.

%%%%%%%%%%%%%%%%%%%%%%%%%%%%%%%%%%%%%%%%%%%%%%%%%%

\subsection{Regression via rank observation}
\label{ssec:method_rank}

\textbf{Rank observation}, or called \emph{pairwise comparison} in the context of regression, indicates the relative order between two real values.
Since the difference of two Gaussian random variables is still Gaussian distributed, the likelihood of the rank can be derived from the cumulative distribution function of a Gaussian distribution:
\begin{equation}
\label{eq:rank}
  Y = T_\mathrm{rank}(Z_1, Z_2) = [Z_1 > Z_2]
,
\quad
  p(Z_1 > Z_2)
=
  \frac12
  \brackets*{
    1 + \erf\parens*{
    \frac{\mu_1 - \mu_2}
    {\sqrt{2(\sigma_1^2 + \sigma_2^2)}}
    }
  }
,
\end{equation}
where $\erf(\cdot)$ is the error function.
Note that even though $\erf(\cdot)$ is a special function, its value can be approximated and its gradient is analytically calculable.\footnote{%
$\erf(x) = \frac{1}{\sqrt{\pi}} \int_{-x}^{x} e^{-t^2} \D{t}$,
and
$\diff{}{x}\erf{(x)} = \frac{2}{\sqrt\pi}e^{-x^2}$
\citep[p.110]{andrews1998special}.
}
Thus the gradient-based optimization is still applicable.

Without loss of generality, we assume $z_1^{(i)} > z_2^{(i)}$ in the dataset.
For a fixed variance $\sigma^2$, the loss function derived from Equations~\eqref{eq:log_likelihood} and \eqref{eq:rank} is
\begin{equation}
  \ell_N^\textrm{rank}(W)
=
  -\frac1N \sum_{i=1}^N
  \log
  \frac12
  \brackets*{
    1 + \erf\parens*{
    \frac{f(x_1^{(i)}; W) - f(x_2^{(i)}; W)}
    {2\sigma}
    }
  }
.
\end{equation}

The rank observation has been studied recently in  \citet{xu2019uncoupled} to solve an uncoupled regression problem, where additional information is required. 
Here, an important question is how good the estimator can be if we only have pairwise comparison.
Intuitively, it should not be consistent.
However, surprisingly, it can be theoretically showed that it is still possible to learn solely from rank observations, and the prediction and the true target differ only by a constant under mild conditions.

\begin{proposition}[Regression via rank observation is consistent up to an additive constant]
\label{prop:rank}
Assuming homoscedastic noise,
$ Z_i = f(X_i; W) + \varepsilon_i $ 
for $ i = 1, \dots, K $, 
where $ \varepsilon_i \iid \Normal(0, \sigma^2) $,
the MLE $\widehat{W}_N$ based on Equation~\eqref{eq:log_likelihood} for rank observations obtained by Equation~\eqref{eq:rank} is consistent up to an additive constant, because
\begin{equation}
  [W_0] 
= 
  \{
  W \in \sW:
  \exists C \in \R,
  f(X; W) - f(X; W_0) = C
  \almosteverywhere
  \} 
.
\end{equation}
\end{proposition}

As a result, we can guarantee that if we know the mean or can obtain a few direct observations, precise prediction becomes possible.
In Section~\ref{ssec:experiment_mean_rank} we validate this finding.
In Appendix~\ref{app:rank}, we provide possible use of the Gumbel distribution, Cauchy distribution, and Exponential distribution for regression via rank observation.

% \red{
% This provides us three ways to use aggregate observations.
% (1) If we do not have any other information, we can analyze the aggregate function $g$ and find the characteristics of the solution, as shown in Propositions~\ref{prop:triplet}, \ref{prop:mean}, and \ref{prop:rank};
% (2) if a small number of direct observations are available, we can naturally combine them with aggregate observations in a ``semi-weakly-supervised'' fashion;
% (3) if we can obtain two kinds of aggregate observations, we can also combine them to maximize the performance.
% Note that it is easy to combine different weak supervisions and also strong supervision within our framework because they are handled based on the same maximum likelihood principle.
% }

\section{Experiments}
\label{sec:experiments}
\begin{table*}
\centering
\caption{
  \textbf{Classification via pairwise similarity and triplet comparison.}
  Means and standard deviations of accuracy (after the optimal permutation) in percentage for $10$ trials are reported.
}
\label{tab:mnist}
\resizebox{\textwidth}{!}{%
\begin{threeparttable}
\begin{tabular}{lcccclcccc}
\toprule
Dataset & 
\multicolumn{1}{c}{Unsupervised} & \multicolumn{3}{c}{Pairwise Similarity} & 
& 
\multicolumn{3}{c}{Triplet Comparison} & \multicolumn{1}{c}{Supervised} 
\\ 
\cmidrule{3-5} 
\cmidrule{7-9} 
& & 
Siamese & Contrastive & Ours\tnote{*} &
& 
Tuplet & Triplet & Ours &  
\\ 
\midrule
MNIST
& $52.30$ & $\mathbf{85.82}$ &$98.45$ &$\mathbf{98.84}$ &
& $18.42$ &$22.77$ &$\mathbf{94.94}$ & $99.04$ 
\\
& $(1.15)$ & $(\mathbf{24.86})$ &$(0.11)$ &$(\mathbf{0.10})$ &
& $(1.08)$ &$(9.38)$ &$(\mathbf{3.68})$ &$(0.08)$ 
\\
FMNIST
& $50.94$ & $62.86$ &$88.49$ &$\mathbf{90.59}$ &
& $21.98$ &$27.27$ &$\mathbf{81.49}$ & $91.97$ 
\\
& $(3.28)$ & $(17.97)$ &$(0.28)$ &$(\mathbf{0.26})$ &
& $(0.72)$ &$(12.82)$ &$(\mathbf{0.94})$ & $(0.24)$ 
\\
KMNIST
& $40.22$ & $61.30$ &$89.65$ &$\mathbf{93.45}$ & 
& $16.00$ &$20.39$ &$\mathbf{81.94}$ & $94.47$ 
\\
& $(0.01)$ & $(17.41)$ &$(0.19)$ &$(\mathbf{0.32})$ &
& $(0.27)$ &$(2.03)$ &$(\mathbf{4.59})$ & $(0.21)$ 
\\
\bottomrule
\end{tabular}
\begin{tablenotes}
  \item[*] See also \citet{hsu2019multiclass}.
  \end{tablenotes}
\end{threeparttable}
} % \resizebox
\end{table*}
\vspace{-1em}

In this section, we present the empirical results of the proposed method.
All experimental details can be found in Appendix~\ref{app:experiment_details}.
More extensive experiments on 20 regression datasets and 30 classification datasets are provided in Appendix~\ref{app:additional_experiments}.
For each type of aggregate observations, outperforming methods are highlighted in boldface using one-sided t-test with a significance level of $5\%$.

%%%%%%%%%%%%%%%%%%%%%%%%%%%%%%%%%%%%%%%%%%%%%%%%%%

\subsection{Classification via triplet comparison}
\label{ssec:experiment_triplet}

We demonstrate that a multiclass classifier is learnable using only triplet comparison data introduced in Section~\ref{ssec:method_triplet}.
We evaluate our method on three image classification datasets, namely
MNIST,\footnote{
  MNIST
  \citep{lecun1998gradient}
  \url{http://yann.lecun.com/exdb/mnist/}
}
Fashion-MNIST (FMNIST),\footnote{
  Fashion-MNIST 
  \citep{xiao2017fashion}
  \url{https://github.com/zalandoresearch/fashion-mnist}
}
and Kuzushiji-MNIST (KMNIST),\footnote{
  Kuzushiji-MNIST
  \citep{clanuwat2018deep}
  \url{http://codh.rois.ac.jp/kmnist/}
}
which consist of $28 \times 28$ grayscale images in $10$ classes.

As for the similarity measure $d$, we followed \citet{cui2020classification} and simply used 
$d = [Z_i \neq Z_j]$ 
as the similarity measure between classes.
We also compared learning from pairwise similarity data, which was studied in \citet{hsu2019multiclass}.
Both pairwise similarity and triplet comparison observations were generated according to our assumptions in Section~\ref{ssec:assumptions}.
Since both learning from pairwise similarities and triplet comparisons are only consistent \emph{up to a permutation} at best, we followed \citet{hsu2019multiclass} and evaluated the performance by modified accuracy that allows any permutation of classes.
The optimal permutation is obtained by solving a linear sum assignment problem using a small amount of individually labeled data \citep{kuhn1955hungarian}.

\textbf{Baseline.}
We used K-means as the unsupervised clustering baseline and the fully supervised learning method as a reference.
We also compared representation learning and metric learning methods such as the Siamese network \citep{koch2015siamese} and contrastive loss \citep{hadsell2006dimensionality} for pairwise similarity, the (2+1)-tuplet loss \citep{sohn2016improved} and the triplet loss \citep{schroff2015facenet} for triplet comparison.
Since the output of such methods is a vector representation, we performed K-means to obtain a prediction so that the results can be directly compared.

\textbf{Results.} Table~\ref{tab:mnist} shows that our method outperforms representation learning methods.
It demonstrates that if the goal is classification, directly optimizing a classification-oriented loss is better than combining representation learning and clustering.
Representation learning from triplet comparison also suffers from a lack of data 
% if the data is i.i.d.~drawn from the distribution.
% This is
because a large amount of triplet comparison data could be in the form of 
``$Z_1$ is equally similar/dissimilar to $Z_2$ and $Z_3$''
if either $Z_2$ and $Z_3$ belong to the same class or $Z_{1:3}$ belong to three different classes.
Such data cannot be used for representation learning but still can provide some information for the classification task.

%%%%%%%%%%%%%%%%%%%%%%%%%%%%%%%%%%%%%%%%%%%%%%%%%%

\begin{table*}
\caption{
\textbf{Regression via mean observation and rank observation on UCI benchmark datasets.} 
Means and standard deviations of error variance (rank observations) or MSE (otherwise) for $10$ trials are reported.
We compare linear regression (LR) and gradient boosting machines (GBM) as the regression function.
}
\label{tab:uci}
\centering
\resizebox{\textwidth}{!}{%
\begin{tabular}{lcccclcccclcc}
\toprule
Dataset & 
\multicolumn{4}{c}{Mean Observation} & 
& 
\multicolumn{4}{c}{Rank Observation} &
&
\multicolumn{2}{c}{Supervised} 
\\ 
\cmidrule{2-5} \cmidrule{7-10}
& 
\multicolumn{2}{c}{Baseline} & 
\multicolumn{2}{c}{Ours} &
& 
\multicolumn{2}{c}{RankNet, Gumbel} &
\multicolumn{2}{c}{Ours, Gaussian} & 
&
\\ 
\cmidrule(lr){2-3} 
\cmidrule(lr){4-5} 
\cmidrule(lr){7-8} 
\cmidrule(lr){9-10} 
\cmidrule(lr){12-13}
 & LR & GBM & LR & GBM & 
 & LR & GBM & LR & GBM & 
 & LR & GBM 
\\ 
\midrule
abalone
& $7.91$ &$7.89$ &$5.27$ &$\mathbf{4.80}$ && $5.81$ &$10.66$ &$\mathbf{5.30}$ &$\mathbf{5.04}$ && $5.00$ & $4.74$
\\
& $(0.4)$ &$(0.5)$ &$(0.4)$ &$(\mathbf{0.3})$ && $(0.4)$ &$(0.7)$ &$(\mathbf{0.3})$ &$(\mathbf{0.5})$ && $(0.3)$ & $(0.4)$
\\
airfoil
& $38.57$ &$28.65$ &$23.59$ &$\mathbf{4.63}$ && $37.15$ &$47.46$ &$27.95$ &$\mathbf{6.18}$ && $22.59$ & $3.84$
\\
& $(2.0)$ &$(2.5)$ &$(1.8)$ &$(\mathbf{0.9})$ && $(1.8)$ &$(3.7)$ &$(1.1)$ &$(\mathbf{1.0})$ && $(1.9)$ & $(0.5)$
\\
auto-mpg
& $41.59$ &$36.31$ &$14.61$ &$\mathbf{9.53}$ && $27.26$ &$65.39$ &$17.34$ &$\mathbf{9.97}$ && $11.73$ & $7.91$
\\
& $(5.7)$ &$(1.9)$ &$(3.2)$ &$(\mathbf{2.4})$ && $(4.0)$ &$(7.4)$ &$(2.0)$ &$(\mathbf{2.0})$ && $(2.3)$ & $(1.6)$
\\
concrete
& $198.51$ &$172.35$ &$115.06$ &$\mathbf{31.84}$ && $244.06$ &$268.86$ &$233.93$ &$\mathbf{38.11}$ && $111.92$ & $24.80$
\\
& $(12.8)$ &$(15.2)$ &$(10.1)$ &$(\mathbf{3.0})$ && $(17.1)$ &$(26.5)$ &$(20.0)$ &$(\mathbf{5.4})$ && $(6.4)$ & $(5.7)$
\\
housing
& $67.40$ &$52.23$ &$27.54$ &$\mathbf{14.85}$ && $52.51$ &$93.07$ &$44.40$ &$\mathbf{23.49}$ && $29.66$ & $13.12$
\\
& $(20.8)$ &$(6.0)$ &$(6.8)$ &$(\mathbf{3.0})$ && $(10.8)$ &$(8.1)$ &$(13.4)$ &$(\mathbf{6.9})$ && $(6.1)$ & $(3.7)$
\\
power-plant
& $172.64$ &$170.10$ &$20.73$ &$\mathbf{12.82}$ && $163.64$ &$294.07$ &$44.82$ &$\mathbf{26.06}$ && $21.17$ & $11.84$
\\
& $(7.1)$ &$(3.4)$ &$(0.8)$ &$(\mathbf{0.6})$ && $(4.8)$ &$(4.9)$ &$(6.1)$ &$(\mathbf{2.5})$ && $(1.0)$ & $(0.9)$
\\
\bottomrule
\end{tabular}
} % \resizebox
\end{table*}

\vspace{-1em}

\subsection{Regression via mean/rank observation}
\label{ssec:experiment_mean_rank}

We compare the performance in regression via direct/mean/rank observations introduced in Sections~\ref{ssec:method_mean} and~\ref{ssec:method_rank}.
We present results on benchmark datasets to show the effectiveness of our method.

\textbf{Baseline.}
For the mean observation, we used a method treating the mean as the true label for each instance as the baseline.
For the rank observation, we used RankNet \citep{burges2005learning} for regression to compare different distribution hypotheses.

%%%%%%%%%%%%%%%%%%%%%%%%%%%%%%%%%%%%%%%%%%%%%%%%%%

\textbf{Real-world dataset.}
We conducted experiments on $6$ UCI benchmark datasets\footnote{
UCI Machine Learning Repository \citep{UCI}
\url{https://archive.ics.uci.edu}
}
and compared linear models and gradient boosting machines as the regression function.
Mean observations of four instances and rank observations are generated according to our assumptions in Section~\ref{ssec:assumptions}.
Since learning from rank observations is only consistent up to an additive constant, we measured the performance by modified MSE that allows any constant shift.
This metric coincides with the variance of the error.
If the estimator is unbiased, it is equal to MSE.
Concretely,
\begin{equation}
  \min_C \frac1N \sum_{i=1}^N 
  \parens*{Z_i - (\widehat{Z}_i + C)}^2
=
  \Var[Z - \widehat{Z}]
.
\end{equation}

\textbf{Results.}
In Table~\ref{tab:uci}, we report error variance for learning from rank observations, otherwise MSE.
It shows that learning from mean observations achieved MSE that is comparable to learning from direct observations, while the error variance of learn from rank observations is slightly higher.
Further, our method consistently outperforms the baseline for the mean observation and RankNet for regression via rank observation.
The performance ranking of learning from direct/mean/rank observations is roughly maintained regardless of the complexity of the model, while the gradient boosting machine usually performs better than the linear model.

\section{Conclusions}
We presented a framework for \emph{learning from aggregate observations}, where only supervision signals given to sets of instances are available.
We proposed a simple yet effective method based on the \emph{maximum likelihood} principle, which can be simply implemented for various differentiable models, including deep neural networks and gradient boosting machines.
We also theoretically analyzed the characteristic of our proposed method based on the concept of consistency up to an equivalent relation.
Experiments on classification via pairwise similarity/triplet comparison and regression via mean/rank observation suggested the feasibility and the usefulness of our method.

\newpage
\section{Broader impact}
In this work we proposed a general method that learns from supervision signals given to sets of instances.
Such studies could provide new tools for privacy preserving machine learning because individual labels are not needed.
This leads to a new way to anonymize data and alleviate potential threats to privacy-sensitive information, in addition to well-known differential privacy techniques \citep{dwork2014algorithmic}, which inject noise into the data to guarantee the anonymity of individuals.

However, such studies may have some negative consequences because depending on the type of aggregate observations, in the worst case, it may be possible to uncover individual information to some extent from aggregated statistics, even if individual labels are not available in the training data.
Nevertheless, a person can deny the uncovered label because of a lack of evidence.
So learning from aggregate observations is still arguably safer than the fully-supervised counterparts in terms of privacy preservation.

Our framework also opens possibilities of using machine learning technology in new problem domains where true labels cannot be straightforwardly obtained, but information such as pairwise/triplet comparisons or coarse-grained data are available or possible to collect.
Finally, we believe that theoretical understanding could provide a better foundation towards solving learning from aggregate observations more effectively in the future.

\section*{Acknowledgement}
We thank Zhenghang Cui, Takuya Shimada, Liyuan Xu, and Zijian Xu for helpful discussion.
We also would like to thank the Supercomputing Division, Information Technology Center, The University of Tokyo, for providing the Reedbush supercomputer system.
NC was supported by MEXT scholarship, JST AIP Challenge, and Google PhD Fellowship program.
MS was supported by JST CREST Grant Number JPMJCR18A2.

\bibliography{references}

\begin{thebibliography}{65}
\providecommand{\natexlab}[1]{#1}
\providecommand{\url}[1]{\texttt{#1}}
\expandafter\ifx\csname urlstyle\endcsname\relax
  \providecommand{\doi}[1]{doi: #1}\else
  \providecommand{\doi}{doi: \begingroup \urlstyle{rm}\Url}\fi

\bibitem[Amar et~al.(2001)Amar, Dooly, Goldman, and Zhang]{amar2001multiple}
Robert~A Amar, Daniel~R Dooly, Sally~A Goldman, and Qi~Zhang.
\newblock Multiple-instance learning of real-valued data.
\newblock In \emph{Proceedings of the Eighteenth International Conference on
  Machine Learning}, ICML '01, pages 3--10, San Francisco, CA, USA, 2001.
  Morgan Kaufmann Publishers Inc.
\newblock ISBN 1-55860-778-1.

\bibitem[Andrews(1998)]{andrews1998special}
Larry~C Andrews.
\newblock \emph{Special functions of mathematics for engineers}, volume~49.
\newblock Spie Press, 1998.

\bibitem[Baghshah and Shouraki(2010)]{baghshah2010kernel}
Mahdieh~Soleymani Baghshah and Saeed~Bagheri Shouraki.
\newblock Kernel-based metric learning for semi-supervised clustering.
\newblock \emph{Neurocomputing}, 73\penalty0 (7-9):\penalty0 1352--1361, 2010.

\bibitem[Bao et~al.(2018)Bao, Niu, and Sugiyama]{bao2018classification}
Han Bao, Gang Niu, and Masashi Sugiyama.
\newblock Classification from pairwise similarity and unlabeled data.
\newblock In Jennifer Dy and Andreas Krause, editors, \emph{Proceedings of the
  35th International Conference on Machine Learning}, volume~80 of
  \emph{Proceedings of Machine Learning Research}, pages 452--461,
  Stockholmsmässan, Stockholm Sweden, 10--15 Jul 2018. PMLR.

\bibitem[Basu et~al.(2004)Basu, Bilenko, and Mooney]{basu2004probabilistic}
Sugato Basu, Mikhail Bilenko, and Raymond~J Mooney.
\newblock A probabilistic framework for semi-supervised clustering.
\newblock In \emph{Proceedings of the tenth ACM SIGKDD international conference
  on Knowledge discovery and data mining}, pages 59--68. ACM, 2004.

\bibitem[Bilenko et~al.(2004)Bilenko, Basu, and Mooney]{bilenko2004integrating}
Mikhail Bilenko, Sugato Basu, and Raymond~J Mooney.
\newblock Integrating constraints and metric learning in semi-supervised
  clustering.
\newblock In \emph{Proceedings of the twenty-first international conference on
  Machine learning}, page~11. ACM, 2004.

\bibitem[Bishop(2006)]{bishop2006pattern}
Christopher~M Bishop.
\newblock \emph{Pattern recognition and machine learning}.
\newblock Springer, 2006.

\bibitem[Bradley and Terry(1952)]{bradley1952rank}
Ralph~Allan Bradley and Milton~E Terry.
\newblock Rank analysis of incomplete block designs: I. the method of paired
  comparisons.
\newblock \emph{Biometrika}, 39\penalty0 (3/4):\penalty0 324--345, 1952.

\bibitem[Burges et~al.(2005)Burges, Shaked, Renshaw, Lazier, Deeds, Hamilton,
  and Hullender]{burges2005learning}
Christopher Burges, Tal Shaked, Erin Renshaw, Ari Lazier, Matt Deeds, Nicole
  Hamilton, and Gregory~N Hullender.
\newblock Learning to rank using gradient descent.
\newblock In \emph{Proceedings of the 22nd International Conference on Machine
  learning}, pages 89--96, 2005.

\bibitem[Cao et~al.(2007)Cao, Qin, Liu, Tsai, and Li]{cao2007learning}
Zhe Cao, Tao Qin, Tie-Yan Liu, Ming-Feng Tsai, and Hang Li.
\newblock Learning to rank: from pairwise approach to listwise approach.
\newblock In \emph{Proceedings of the 24th international conference on Machine
  learning}, pages 129--136, 2007.

\bibitem[Carbonneau et~al.(2018)Carbonneau, Cheplygina, Granger, and
  Gagnon]{carbonneau2018multiple}
Marc-Andr{\'e} Carbonneau, Veronika Cheplygina, Eric Granger, and Ghyslain
  Gagnon.
\newblock Multiple instance learning: A survey of problem characteristics and
  applications.
\newblock \emph{Pattern Recognition}, 77:\penalty0 329--353, 2018.

\bibitem[Chang and Lin(2011)]{libsvm}
Chih-Chung Chang and Chih-Jen Lin.
\newblock {LIBSVM}: a library for support vector machines.
\newblock \emph{ACM transactions on intelligent systems and technology (TIST)},
  2\penalty0 (3):\penalty0 27, 2011.

\bibitem[Clanuwat et~al.(2018)Clanuwat, Bober-Irizar, Kitamoto, Lamb, Yamamoto,
  and Ha]{clanuwat2018deep}
Tarin Clanuwat, Mikel Bober-Irizar, Asanobu Kitamoto, Alex Lamb, Kazuaki
  Yamamoto, and David Ha.
\newblock Deep learning for classical japanese literature, 2018.

\bibitem[Cui et~al.(2020)Cui, Charoenphakdee, Sato, and
  Sugiyama]{cui2020classification}
Zhenghang Cui, Nontawat Charoenphakdee, Issei Sato, and Masashi Sugiyama.
\newblock Classification from triplet comparison data.
\newblock \emph{Neural Computation}, 32\penalty0 (3):\penalty0 659--681, 2020.

\bibitem[Davis et~al.(2007)Davis, Kulis, Jain, Sra, and
  Dhillon]{davis2007information}
Jason~V Davis, Brian Kulis, Prateek Jain, Suvrit Sra, and Inderjit~S Dhillon.
\newblock Information-theoretic metric learning.
\newblock In \emph{Proceedings of the 24th international conference on Machine
  learning}, pages 209--216. ACM, 2007.

\bibitem[Dietterich et~al.(1997)Dietterich, Lathrop, and
  Lozano-P{\'e}rez]{dietterich1997solving}
Thomas~G Dietterich, Richard~H Lathrop, and Tom{\'a}s Lozano-P{\'e}rez.
\newblock Solving the multiple instance problem with axis-parallel rectangles.
\newblock \emph{Artificial intelligence}, 89\penalty0 (1-2):\penalty0 31--71,
  1997.

\bibitem[Dua and Graff(2017)]{UCI}
Dheeru Dua and Casey Graff.
\newblock {UCI} machine learning repository, 2017.
\newblock URL \url{http://archive.ics.uci.edu/ml}.

\bibitem[Dwork et~al.(2014)Dwork, Roth, et~al.]{dwork2014algorithmic}
Cynthia Dwork, Aaron Roth, et~al.
\newblock The algorithmic foundations of differential privacy.
\newblock \emph{Foundations and Trends{\textregistered} in Theoretical Computer
  Science}, 9\penalty0 (3--4):\penalty0 211--407, 2014.

\bibitem[Flaxman et~al.(2015)Flaxman, Wang, and Smola]{flaxman2015supported}
Seth~R Flaxman, Yu-Xiang Wang, and Alexander~J Smola.
\newblock Who supported {Obama} in 2012? ecological inference through
  distribution regression.
\newblock In \emph{Proceedings of the 21th ACM SIGKDD International Conference
  on Knowledge Discovery and Data Mining}, pages 289--298, 2015.

\bibitem[Friedman(2001)]{friedman2001greedy}
Jerome~H Friedman.
\newblock Greedy function approximation: a gradient boosting machine.
\newblock \emph{Annals of statistics}, pages 1189--1232, 2001.

\bibitem[Gutierrez et~al.(2015)Gutierrez, Perez-Ortiz, Sanchez-Monedero,
  Fernandez-Navarro, and Hervas-Martinez]{gutierrez2015ordinal}
Pedro~Antonio Gutierrez, Maria Perez-Ortiz, Javier Sanchez-Monedero, Francisco
  Fernandez-Navarro, and Cesar Hervas-Martinez.
\newblock Ordinal regression methods: survey and experimental study.
\newblock \emph{IEEE Transactions on Knowledge and Data Engineering},
  28\penalty0 (1):\penalty0 127--146, 2015.

\bibitem[Hadsell et~al.(2006)Hadsell, Chopra, and
  LeCun]{hadsell2006dimensionality}
Raia Hadsell, Sumit Chopra, and Yann LeCun.
\newblock Dimensionality reduction by learning an invariant mapping.
\newblock In \emph{2006 IEEE Computer Society Conference on Computer Vision and
  Pattern Recognition (CVPR'06)}, volume~2, pages 1735--1742. IEEE, 2006.

\bibitem[Haghiri et~al.(2018)Haghiri, Garreau, and von
  Luxburg]{haghiri2018comparison}
Siavash Haghiri, Damien Garreau, and Ulrike von Luxburg.
\newblock Comparison-based random forests.
\newblock In \emph{Proceedings of the 35th International Conference on Machine
  Learning}, pages 1871--1880, 2018.

\bibitem[Horvitz and Mulligan(2015)]{horvitz2015data}
Eric Horvitz and Deirdre Mulligan.
\newblock Data, privacy, and the greater good.
\newblock \emph{Science}, 349\penalty0 (6245):\penalty0 253--255, 2015.

\bibitem[Hsu et~al.(2019)Hsu, Lv, Schlosser, Odom, and Kira]{hsu2019multiclass}
Yen-Chang Hsu, Zhaoyang Lv, Joel Schlosser, Phillip Odom, and Zsolt Kira.
\newblock Multi-class classification without multi-class labels.
\newblock In \emph{International Conference on Learning Representations}, 2019.

\bibitem[Jordan and Mitchell(2015)]{jordan2015machine}
Michael~I Jordan and Tom~M Mitchell.
\newblock Machine learning: Trends, perspectives, and prospects.
\newblock \emph{Science}, 349\penalty0 (6245):\penalty0 255--260, 2015.

\bibitem[Ke et~al.(2017)Ke, Meng, Finley, Wang, Chen, Ma, Ye, and
  Liu]{ke2017lightgbm}
Guolin Ke, Qi~Meng, Thomas Finley, Taifeng Wang, Wei Chen, Weidong Ma, Qiwei
  Ye, and Tie-Yan Liu.
\newblock {LightGBM}: A highly efficient gradient boosting decision tree.
\newblock In \emph{Advances in neural information processing systems}, pages
  3146--3154, 2017.

\bibitem[King et~al.(2004)King, Tanner, and Rosen]{king2004ecological}
Gary King, Martin~A Tanner, and Ori Rosen.
\newblock \emph{Ecological inference: New methodological strategies}.
\newblock Cambridge University Press, 2004.

\bibitem[Koch et~al.(2015)Koch, Zemel, and Salakhutdinov]{koch2015siamese}
Gregory Koch, Richard Zemel, and Ruslan Salakhutdinov.
\newblock Siamese neural networks for one-shot image recognition.
\newblock In \emph{ICML deep learning workshop}, volume~2. Lille, 2015.

\bibitem[K\"{u}ck and de~Freitas(2005)]{kuck2012learning}
Hendrik K\"{u}ck and Nando de~Freitas.
\newblock Learning about individuals from group statistics.
\newblock In \emph{Proceedings of the Twenty-First Conference on Uncertainty in
  Artificial Intelligence}, UAI'05, 2005.

\bibitem[Kuhn(1955)]{kuhn1955hungarian}
Harold~W Kuhn.
\newblock The hungarian method for the assignment problem.
\newblock \emph{Naval research logistics quarterly}, 2\penalty0 (1-2):\penalty0
  83--97, 1955.

\bibitem[Kumar and Kummamuru(2008)]{kumar2008semisupervised}
Nimit Kumar and Krishna Kummamuru.
\newblock Semisupervised clustering with metric learning using relative
  comparisons.
\newblock \emph{IEEE Transactions on Knowledge and Data Engineering},
  20\penalty0 (4):\penalty0 496--503, 2008.

\bibitem[Law et~al.(2018)Law, Sejdinovic, Cameron, Lucas, Flaxman, Battle, and
  Fukumizu]{law2018variational}
Ho~Chung Law, Dino Sejdinovic, Ewan Cameron, Tim Lucas, Seth Flaxman, Katherine
  Battle, and Kenji Fukumizu.
\newblock Variational learning on aggregate outputs with gaussian processes.
\newblock In \emph{Advances in Neural Information Processing Systems}, pages
  6081--6091, 2018.

\bibitem[LeCun et~al.(1998)LeCun, Bottou, Bengio, Haffner,
  et~al.]{lecun1998gradient}
Yann LeCun, L{\'e}on Bottou, Yoshua Bengio, Patrick Haffner, et~al.
\newblock Gradient-based learning applied to document recognition.
\newblock \emph{Proceedings of the IEEE}, 86\penalty0 (11):\penalty0
  2278--2324, 1998.

\bibitem[LeCun et~al.(2015)LeCun, Bengio, and Hinton]{lecun2015deep}
Yann LeCun, Yoshua Bengio, and Geoffrey Hinton.
\newblock Deep learning.
\newblock \emph{nature}, 521\penalty0 (7553):\penalty0 436, 2015.

\bibitem[Liu and Tao(2014)]{liu2014robustness}
Tongliang Liu and Dacheng Tao.
\newblock On the robustness and generalization of cauchy regression.
\newblock In \emph{2014 4th IEEE International Conference on Information
  Science and Technology}, pages 100--105. IEEE, 2014.

\bibitem[Loshchilov and Hutter(2019)]{loshchilov2018decoupled}
Ilya Loshchilov and Frank Hutter.
\newblock Decoupled weight decay regularization.
\newblock In \emph{International Conference on Learning Representations}, 2019.

\bibitem[Luce(1959)]{luce1959individual}
R~Duncan Luce.
\newblock \emph{Individual choice behavior}.
\newblock John Wiley, 1959.

\bibitem[Mojsilovic and Ukkonen(2019)]{mojsilovic2019relative}
Stefan Mojsilovic and Antti Ukkonen.
\newblock Relative distance comparisons with confidence judgements.
\newblock In \emph{Proceedings of the 2019 SIAM International Conference on
  Data Mining}, pages 459--467. SIAM, 2019.

\bibitem[Patrini et~al.(2014)Patrini, Nock, Rivera, and
  Caetano]{patrini2014almost}
Giorgio Patrini, Richard Nock, Paul Rivera, and Tiberio Caetano.
\newblock {(Almost)} no label no cry.
\newblock In \emph{Advances in Neural Information Processing Systems}, pages
  190--198, 2014.

\bibitem[Patrini et~al.(2017)Patrini, Rozza, Krishna~Menon, Nock, and
  Qu]{patrini2017making}
Giorgio Patrini, Alessandro Rozza, Aditya Krishna~Menon, Richard Nock, and
  Lizhen Qu.
\newblock Making deep neural networks robust to label noise: A loss correction
  approach.
\newblock In \emph{Proceedings of the IEEE Conference on Computer Vision and
  Pattern Recognition}, pages 1944--1952, 2017.

\bibitem[Perrot and Von~Luxburg(2019)]{perrot2019boosting}
Micha{\"e}l Perrot and Ulrike Von~Luxburg.
\newblock Boosting for comparison-based learning.
\newblock In \emph{Proceedings of the Twenty-Eighth International Joint
  Conference on Artificial Intelligence, {IJCAI} 2019, Macao, China, August
  10-16, 2019}, pages 1844--1850, 2019.

\bibitem[Plackett(1975)]{plackett1975analysis}
Robin~L Plackett.
\newblock The analysis of permutations.
\newblock \emph{Journal of the Royal Statistical Society: Series C (Applied
  Statistics)}, 24\penalty0 (2):\penalty0 193--202, 1975.

\bibitem[Quadrianto et~al.(2009)Quadrianto, Smola, Caetano, and
  Le]{quadrianto2009estimating}
Novi Quadrianto, Alex~J Smola, Tiberio~S Caetano, and Quoc~V Le.
\newblock Estimating labels from label proportions.
\newblock \emph{Journal of Machine Learning Research}, 10\penalty0
  (Oct):\penalty0 2349--2374, 2009.

\bibitem[Ray and Page(2001)]{ray2001multiple}
Soumya Ray and David Page.
\newblock Multiple instance regression.
\newblock In \emph{Proceedings of the Eighteenth International Conference on
  Machine Learning}, ICML '01, pages 425--432, San Francisco, CA, USA, 2001.
  Morgan Kaufmann Publishers Inc.
\newblock ISBN 1-55860-778-1.

\bibitem[Schroff et~al.(2015)Schroff, Kalenichenko, and
  Philbin]{schroff2015facenet}
Florian Schroff, Dmitry Kalenichenko, and James Philbin.
\newblock {FaceNet}: A unified embedding for face recognition and clustering.
\newblock In \emph{Proceedings of the IEEE conference on computer vision and
  pattern recognition}, pages 815--823, 2015.

\bibitem[Schuessler(1999)]{schuessler1999ecological}
Alexander~A Schuessler.
\newblock Ecological inference.
\newblock \emph{Proceedings of the National Academy of Sciences}, 96\penalty0
  (19):\penalty0 10578--10581, 1999.

\bibitem[Schultz and Joachims(2004)]{schultz2004learning}
Matthew Schultz and Thorsten Joachims.
\newblock Learning a distance metric from relative comparisons.
\newblock In \emph{Advances in neural information processing systems}, pages
  41--48, 2004.

\bibitem[Sheldon and Dietterich(2011)]{sheldon2011collective}
Daniel~R Sheldon and Thomas~G Dietterich.
\newblock Collective graphical models.
\newblock In \emph{Advances in Neural Information Processing Systems}, pages
  1161--1169, 2011.

\bibitem[Sohn(2016)]{sohn2016improved}
Kihyuk Sohn.
\newblock Improved deep metric learning with multi-class n-pair loss objective.
\newblock In \emph{Advances in neural information processing systems}, pages
  1857--1865, 2016.

\bibitem[Tamuz et~al.(2011)Tamuz, Liu, Belongie, Shamir, and
  Kalai]{tamuz2011adaptively}
Omer Tamuz, Ce~Liu, Serge Belongie, Ohad Shamir, and Adam~Tauman Kalai.
\newblock Adaptively learning the crowd kernel.
\newblock \emph{ICML}, 2011.

\bibitem[Tanaka et~al.(2019{\natexlab{a}})Tanaka, Iwata, Tanaka, Kurashima,
  Okawa, and Toda]{tanaka2019refining}
Yusuke Tanaka, Tomoharu Iwata, Toshiyuki Tanaka, Takeshi Kurashima, Maya Okawa,
  and Hiroyuki Toda.
\newblock Refining coarse-grained spatial data using auxiliary spatial data
  sets with various granularities.
\newblock In \emph{Proceedings of the AAAI Conference on Artificial
  Intelligence}, volume~33, pages 5091--5099, 2019{\natexlab{a}}.

\bibitem[Tanaka et~al.(2019{\natexlab{b}})Tanaka, Tanaka, Iwata, Kurashima,
  Okawa, Akagi, and Toda]{tanaka2019spatially}
Yusuke Tanaka, Toshiyuki Tanaka, Tomoharu Iwata, Takeshi Kurashima, Maya Okawa,
  Yasunori Akagi, and Hiroyuki Toda.
\newblock Spatially aggregated gaussian processes with multivariate areal
  outputs.
\newblock In \emph{Advances in Neural Information Processing Systems}, pages
  3000--3010, 2019{\natexlab{b}}.

\bibitem[Thurstone(1927)]{thurstone1927law}
Louis~L Thurstone.
\newblock A law of comparative judgment.
\newblock \emph{Psychological review}, 34\penalty0 (4):\penalty0 273, 1927.

\bibitem[Van~der Vaart(2000)]{van2000asymptotic}
Aad~W Van~der Vaart.
\newblock \emph{Asymptotic statistics}, volume~3.
\newblock Cambridge university press, 2000.

\bibitem[Vanschoren et~al.(2013)Vanschoren, van Rijn, Bischl, and
  Torgo]{OpenML2013}
Joaquin Vanschoren, Jan~N. van Rijn, Bernd Bischl, and Luis Torgo.
\newblock {OpenML}: Networked science in machine learning.
\newblock \emph{SIGKDD Explorations}, 15\penalty0 (2):\penalty0 49--60, 2013.
\newblock \doi{10.1145/2641190.2641198}.
\newblock URL \url{http://doi.acm.org/10.1145/2641190.2641198}.

\bibitem[Wald(1949)]{wald1949note}
Abraham Wald.
\newblock Note on the consistency of the maximum likelihood estimate.
\newblock \emph{The Annals of Mathematical Statistics}, 20\penalty0
  (4):\penalty0 595--601, 1949.

\bibitem[Xiao et~al.(2017)Xiao, Rasul, and Vollgraf]{xiao2017fashion}
Han Xiao, Kashif Rasul, and Roland Vollgraf.
\newblock {Fashion-MNIST}: a novel image dataset for benchmarking machine
  learning algorithms, 2017.

\bibitem[Xing et~al.(2003)Xing, Jordan, Russell, and Ng]{xing2003distance}
Eric~P Xing, Michael~I Jordan, Stuart~J Russell, and Andrew~Y Ng.
\newblock Distance metric learning with application to clustering with
  side-information.
\newblock In \emph{Advances in neural information processing systems}, pages
  521--528, 2003.

\bibitem[Xu et~al.(2019)Xu, Honda, Niu, and Sugiyama]{xu2019uncoupled}
Liyuan Xu, Junya Honda, Gang Niu, and Masashi Sugiyama.
\newblock Uncoupled regression from pairwise comparison data.
\newblock In \emph{Advances in Neural Information Processing Systems 32}, pages
  3994--4004. Curran Associates, Inc., 2019.

\bibitem[Yang(2005)]{yang2005review}
Jun Yang.
\newblock Review of multi-instance learning and its applications.
\newblock \emph{Technical report, School of Computer Science Carnegie Mellon
  University}, 2005.

\bibitem[Yu et~al.(2013)Yu, Liu, Kumar, Tony, and Chang]{felix2013psvm}
Felix Yu, Dong Liu, Sanjiv Kumar, Jebara Tony, and Shih-Fu Chang.
\newblock {$\propto$SVM} for learning with label proportions.
\newblock In \emph{Proceedings of the 30th International Conference on Machine
  Learning}, pages 504--512, 2013.

\bibitem[Zhang et~al.(2019)Zhang, Charoenphakdee, and
  Sugiyama]{zhang2019learning}
Yivan Zhang, Nontawat Charoenphakdee, and Masashi Sugiyama.
\newblock Learning from indirect observations.
\newblock \emph{arXiv preprint arXiv:1910.04394}, 2019.

\bibitem[Zhou(2004)]{zhou2004multi}
Zhi-Hua Zhou.
\newblock Multi-instance learning: A survey.
\newblock \emph{Department of Computer Science \& Technology, Nanjing
  University, Tech. Rep}, 2004.

\bibitem[Zhou(2017)]{zhou2017brief}
Zhi-Hua Zhou.
\newblock A brief introduction to weakly supervised learning.
\newblock \emph{National Science Review}, 5\penalty0 (1):\penalty0 44--53,
  2017.

\end{thebibliography}

\newpage
\appendix
\part*{\Large{Appendix}}
\everymath{\displaystyle}

%%%%%%%%%%%%%%%%%%%%%%%%%%%%%%%%%%%%%%%%%%%%%%%%%%

\section{Proofs}
\label{app:proofs}

In this section, we provide missing proofs of Proposition~\ref{prop:triplet}, Proposition~\ref{prop:mean}, and Proposition~\ref{prop:rank}.

%%%%%%%%%%%%%%%%%%%%%%%%%%%%%%%%%%%%%%%%%%%%%%%%%%

\subsection{Proof of Proposition~\ref{prop:triplet}}

\begin{proof}

Denote $X' = (X_1, X_2, X_3)$ 
where $X_1, X_2, X_3$ are i.i.d.~observations. 
Let $p_i(X', W) = p(Y = i | X_1, X_2, X_3; W)$, $i = 0, 1$.

Following simple algebra,
\begin{equation}
  \ell(W)
=  
  \E\brackets*{
  \E\brackets*{
  \log p(Y | X', W) | X'
  }}
=
  \E\brackets*{
  \sum_{i=0}^1 p_i(X', W_0) \log p_i(X', W)
  }
.
\end{equation}

So, the difference of $\ell(W)$ and $\ell(W_0)$ satisfies
\begin{equation}
\begin{aligned}
  \ell(W) - \ell(W_0) 
&= 
  \E\brackets*{
  \sum_{i=0}^1 
  p_i(X', W_0)
  \log \dfrac{p_i(X', W)}{p_i(X', W_0)}
  }
\\
&\leq
  \E\brackets*{
  \sum_{i=0}^1
  p_i(X', W_0)
  \parens*{\dfrac{p_i(X', W)}{p_i(X', W_0)} - 1}
  }
\\
&=
  \E\brackets*{
  \sum_{i=0}^1 p_i(X', W) - \sum_{i=0}^1 p_i(X', W_0)
  }
\\
&= 
  0
.
\end{aligned}
\end{equation}

The inequality holds because $\log x \leq x - 1$ for $x > 0$, 
where ``$=$'' holds if and only if $x = 1$.
Hence $\ell(W) = \ell(W_0)$, 
which implies that 
$p_i(X', W) = p_i(X', W_0)$, a.e.~for $i = 0, 1$. 
This is equivalent to $p_1(X', W) = p_1(X', W_0)$.

Let $\sigma$ be a permutation function of 
$\{1, 2, \ldots, C\}$. 
That is, 
$\{\sigma(1), \sigma(2), \ldots, \sigma(C)\} = \{1, 2, \ldots, K\}$. 
If $p(Z = i | X; W_0) = p(Z = \sigma(i) | X;W)$, 
$\forall 1 \leq i \leq K$. 
With the assumption $d(\sigma(i), \sigma(j)) = d(i,j)$, 
we have 
\begin{align*}
  p_1(X', W) 
&=
  \sum_{
  \substack{d(i, j) < d(i, k)\\ 
  i, j, k \in \{1, \dots, C\}}}
  p(Z_1 = i | X_1; W) 
  p(Z_2 = j | X_2; W) 
  p(Z_3 = k | X_3; W)
\\
&=
  \sum_{
  \substack{d(\sigma(i), \sigma(j)) < d(\sigma(i), \sigma(k))\\ 
  \sigma(i), \sigma(j), \sigma(k) \in \{1, \dots, C\}}}
  p(Z_1 = \sigma(i) | X_1; W) 
  p(Z_2 = \sigma(j) | X_2; W) 
  p(Z_3 = \sigma(k) | X_3; W)
\\
&=
  \sum_{
  \substack{d(\sigma(i), \sigma(j)) < d(\sigma(i), \sigma(k))\\ 
  \sigma(i), \sigma(j), \sigma(k) \in \{1, \dots, C\}}}
  p(Z_1 = i | X_1; W) 
  p(Z_2 = j | X_2; W) 
  p(Z_3 = k | X_3; W)
\\
&=
  \sum_{
  \substack{d(i,j)<d(i,k)\\ 
  \sigma(i), \sigma(j), \sigma(k) \in \{1, \dots, C\}}}
  p(Z_1 = i | X_1; W) 
  p(Z_2 = j | X_2; W) 
  p(Z_3 = k | X_3; W)
\\
&=
  \sum_{
  \substack{d(i,j)<d(i,k)\\ 
  i, j, k \in \{1, \dots, C\}}}
  p(Z_1 = i | X_1; W) 
  p(Z_2 = j | X_2; W) 
  p(Z_3 = k | X_3; W) 
= p_1(X', W_0)
.
\end{align*}
On the other hand, if we consider a special case 
where $C = 3$, 
distance function $d(i, j) = [i \neq j]$, 
and 
$
  \parens*{p(Z = 1 | X; W), p(Z = 2 | X; W),p(Z = 3 | X; W)} 
= \parens*{\frac{1}{3} - W, \frac{1}{3}, \frac{1}{3} + W}
$,
then the following equation always holds:
\begin{align*}
  p_1(X', W)
&=
  \sum_{
  \substack{i=j, i\neq k\\ 
  i, j, k \in \{1, 2, 3\}}}
  p(Z_1 = i | X_1; W) 
  p(Z_2 = j | X_2; W) 
  p(Z_3 = k | X_3; W)
\\
&=
  \sum_{i=1}^3\sum_{k\neq i}
  p(Z_1 = i | X_1; W) 
  p(Z_2 = i | X_2; W) 
  p(Z_3 = k | X_3; W)
\\
&=
  \sum_{i=1}^3
  p(Z_1 = i | X_1; W) 
  p(Z_2 = i | X_2; W) 
  (1 - p(Z_3 = i | X_3; W))
\\
&=
  \parens*{\frac{1}{3}-W}^2\parens*{\frac{2}{3}+W}
+ \parens*{\frac{1}{3}}^2\parens*{\frac{2}{3}}
+ \parens*{\frac{1}{3}+W}^2\parens*{\frac{2}{3}-W}
\\
&=
  \frac{2}{9}
.
\end{align*}
In this example, $\forall W_0 \in [0, \frac{1}{3}]$, 
equivalence up to a permutation requires $W = W_0$ or $W = \frac{2}{3} - W_0$ but $[W_0] = [0, \frac{1}{3}]$. 
Hence, this counterexample shows that little information can be learned from triplet comparison if the distance function or the distribution of $Z$ is not modeled properly.

\end{proof}

Further, we can also prove that classification via pairwise similarity is at most consistent up to a permutation, which is empirically shown in \cite{hsu2019multiclass}.
More concretely, we can show that it is consistent up to rotations and reflections in the parameter space.

\begin{proof}
Let $p(Z_i | X_i) \in \Delta^{C-1}$, $i = 1, 2$, be the vector of probability parameters of a categorical distribution, where $\Delta^{C-1}$ denotes $C-1$-dimensional simplex. 
Then the probability of similarity $Y = [Z_1 = Z_2]$, i.e., $X_1$ and $X_2$ belong to the same class, can be written as
\begin{equation}
  p(Y = 1 | X_1, X_2) 
= p(Z_1 | X_1)\T p(Z_2 | X_2)
.
\end{equation}
From $p(Y = 1 | X_1, X_2)$ we cannot identify $p(Z_1 | X_1)$ and $p(Z_2 | X_2)$, because for any orthogonal matrix $Q$, representing rotations and reflections in the parameter space,
\begin{equation}
  (Q p(Z_1 | X_1))\T (Qp(Z_2 | X_2)) 
= p(Z_1 | X_1)\T p(Z_2 | X_2)
\end{equation}
holds for any $p(Z_1 | X_1)$ and $p(Z_2 | X_2)$.
The choice of $Q$ is restricted because $Qp(Z_1 | X_1)$ and $Qp(Z_2 | Z_2)$ should be on the simplex $\Delta^{C-1}$ too. 
Any permutation matrix $P$ is an orthogonal matrix and always satisfies this condition.
So classification via pairwise similarity is at most consistent up to a permutation.

\end{proof}

%%%%%%%%%%%%%%%%%%%%%%%%%%%%%%%%%%%%%%%%%%%%%%%%%%

\subsection{Proof of Proposition~\ref{prop:mean}}

\begin{proof}

Given $X_{1:K}$, 
$
  Y 
= 
  \frac{1}{K}\sum_{i=1}^K Z_i
\sim
  \Normal\parens*{
  \frac{1}{K} \sum_{i=1}^K f(X_i; W_0), 
  \frac{\sigma^2}{K}
  }
$.

Denote 
$ 
  \mu(X_{1:K}, W) 
= 
  \frac{1}{K} \sum_{i=1}^K f(X_i; W)
$. 

We have
\begin{equation}
  \ell(W)
= \E\brackets*{
  \E\brackets*{
  \log p(Y | X_{1:K}; W) | X_{1:K}
  }}
=
- \frac{K}{2\sigma^2}
  \E\brackets*{
  \E\brackets*{
  {(Y - \mu(X_{1:K}, W))^2} | X_{1:K}
  }}
+ C
,
\end{equation}
where $C$ is a constant. 

Noting that
\begin{equation}
\begin{aligned}
&
  \E\brackets*{{(Y - \mu(X_{1:K}, W)^2} | X_{1:K}}
\\
=& 
  \E\brackets*{
  (Y - \mu(X_{1:K}, W_0)
+ \mu(X_{1:K}, W_0) 
- \mu(X_{1:K}, W))^2 | X_{1:K}
  }
\\
=&
  \E\brackets*{(Y - \mu(X_{1:K}, W_0))^2 | X_{1:K}}
+ \E\brackets*{
  (\mu(X_{1:K}, W_0) - \mu(X_{1:K}, W))^2 | X_{1:K}
  }
\\
+&
  2\E\brackets*{
  (Y - \mu(X_{1:K}, W_0))(\mu(X_{1:K}, W_0)
  - \mu(X_{1:K}, W)) | X_{1:K}
  }
\\
=&
  \E\brackets*{(Y - \mu(X_{1:K}, W_0))^2 | X_{1:K}}
+ (\mu(X_{1:K}, W_0) - \mu(X_{1:K}, W))^2
\\
+&
  2(\mu(X_{1:K}, W_0) - \mu(X_{1:K}, W))
  \E\brackets*{Y - \mu(X_{1:K}, W_0) | X_{1:K}}
\\
=&
  \E\brackets*{
  (Y - \mu(X_{1:K}, W_0))^2 | X_{1:K}
  } 
+ (\mu(X_{1:K}, W_0) - \mu(X_{1:K}, W))^2
,
\end{aligned}
\end{equation}

$ \ell(W) = \ell(W_0) $ implies that
\begin{equation}
\label{eq:zero_expectation}
  \E\brackets*{
  (\sum_{i=1}^K f(X_i; W) 
  -\sum_{i=1}^K f(X_i; W_0))^2
  }
= 0
.
\end{equation}

Denote $f(X; W) - f(X; W_0)$ as $\Delta(X, W)$. Equation~\eqref{eq:zero_expectation} can be expanded as follows
\begin{equation}
\begin{aligned}
  0 
&=
  \E\brackets*{(\sum_{i=1}^K \Delta(X_i, W))^2}
\\
&=
  \sum_{i=1}^K
  \E\brackets*{\Delta^2(X_i, W)} 
+ 2\sum_{1\leq i<j\leq K}
  \E\brackets*{\Delta(X_i, W) \Delta(X_j, W)}
\\
&=
  \sum_{i=1}^K
  \E\brackets*{\Delta^2(X_i, W)} 
+ 2\sum_{1\leq i<j\leq K} 
  \E\brackets*{\Delta(X_i, W)}
  \E\brackets*{\Delta(X_j, W)}
\\
&=
  \sum_{i=1}^K
  \E\brackets*{\Delta^2(X_i, W)} + 
  2\sum_{1\leq i<j\leq K} \parens*{\E\brackets*{\Delta(X_i, W)}}^2
.
\end{aligned}
\end{equation}

Thus, 
$
  \E\brackets*{\Delta^2(X_i, W)}
=
  \E\brackets*{\Delta(X_i, W)}
= 
  0
$. 
We have $ \Delta(X_i, W) = 0 $, a.e., 
therefore $ f(X, W) = f(X, W_0) $, a.e.

\end{proof}

%%%%%%%%%%%%%%%%%%%%%%%%%%%%%%%%%%%%%%%%%%%%%%%%%%

\subsection{Proof of Proposition~\ref{prop:rank}}

\begin{proof}

Denote $ X' = (X_1, X_2) $ 
and denote $ p(Y = i | X_1, X_2; W) $ 
as $ p_i(X', W), i = 0, 1 $.

Following simple algebra,
\begin{equation}
  \ell(W)
=  
  \E\brackets*{
  \E\brackets*{
  \log p(Y | X', W) | X'
  }}
=
  \E\brackets*{
  \sum_{i=0}^1 p_i(X', W_0) \log p_i(X', W)
  }
.
\end{equation}

So, the difference of $\ell(W)$ and $\ell(W_0)$ satisfies
\begin{equation}
\begin{aligned}
  \ell(W) - \ell(W_0) 
&= 
  \E\brackets*{
  \sum_{i=0}^1 
  p_i(X', W_0)
  \log \dfrac{p_i(X', W)}{p_i(X', W_0)}
  }
\\
&\leq
  \E\brackets*{
  \sum_{i=0}^1
  p_i(X', W_0)
  \parens*{\dfrac{p_i(X', W)}{p_i(X', W_0)} - 1}
  }
\\
&=
  \E\brackets*{
  \sum_{i=0}^1 p_i(X', W) - \sum_{i=0}^1 p_i(X', W_0)
  }
\\
&= 
  0
.
\end{aligned}
\end{equation}

The inequality holds because $ \log x \leq x - 1 $ for $ x > 0 $, 
where ``$=$'' holds if and only if $ x = 1 $.
Hence $ \ell(W) = \ell(W_0) $, 
which implies that 
$ p_i(X', W) = p_i(X', W_0) $, a.e.~for $ i = 0, 1 $. Together with Equation~\eqref{eq:rank}, we have
\begin{equation}
  \erf\parens*{
  \frac{1}{2\sigma}
  \parens*{f(X_1, W) - f(X_2, W)}
  }
=
  \erf\parens*{
  \frac{1}{2\sigma}
  \parens*{f(X_1, W_0) - f(X_2, W_0)}
  }
.
\end{equation}
Hence, 
$ 
  f(X_1, W) - f(X_2, W) 
= 
  f(X_1, W_0) - f(X_2, W_0) 
$, 
a.e.

\end{proof}

%%%%%%%%%%%%%%%%%%%%%%%%%%%%%%%%%%%%%%%%%%%%%%%%%%

\section{Preliminary: mean squared error and Gaussian distribution}
\label{app:mse}

In this section, we provide preliminaries related to Section~\ref{ssec:method_mean} and Section~\ref{ssec:method_rank}.

\subsection{Mean squared error}
It is well known that the \emph{mean squared error} (MSE) loss function can be derived from the maximum likelihood estimation under an \emph{additive homoscedastic Gaussian noise} model \citep[see also][p.~140]{bishop2006pattern}.
Concretely, let the target $Z$ be given by a deterministic function $f(X; W)$ with additive Gaussian noise $\varepsilon$ with zero mean and fixed variance:
\begin{equation}
  Z = f(X; W) + \varepsilon
\text{, where }
  \varepsilon \sim \Normal(0, \sigma^2)
.
\end{equation}
Or equivalently, 
\begin{equation}
\label{eq:gaussian_noise}
  Z \sim \Normal(\mu, \sigma^2)
\text{, where }
  \mu = f(X; W)
.
\end{equation}
Here we assume the noise is \emph{homoscedastic} so the variance of the noise $\sigma^2$ is a fixed value, i.e., the noise is independent on the values of features $X$.
Thus the parameter is only the mean, $\theta = \{\mu\}$.
This assumption may be inappropriate for some applications when the noise is \emph{heteroscedastic}.
In such case, we can let the deterministic function $f$ predict both the mean $\mu$ and the variance $\sigma^2$, i.e., $\theta = \{\mu, \sigma^2\}$.

The log-likelihood derived from Equation~\eqref{eq:gaussian_noise} is
\begin{equation}
\label{eq:gaussian_likelihood}
  \log p(\{x^{(i)}, z^{(i)}\}_{i=1}^n; W)
=
  -\frac1n \sum_{i=1}^n
  \frac{\parens*{z^{(i)} - f(x^{(i)}; W)}^2}{2\sigma^2}
- \frac12\log(2\pi\sigma^2)
.
\end{equation}

To maximize the log-likelihood w.r.t.~$W$, the loss function can be defined by
\begin{equation}
\label{eq:mse}
  L(W)
=
  \frac1n \sum_{i=1}^n
  \parens*{z^{(i)} - f(x^{(i)}; W)}^2
,
\end{equation}
which is the mean squared error (MSE).
Denote the maximum likelihood estimator of $W$ by $ \widehat{W}_{ML} = \argmin_W L(W) $.

Note that the variance $\sigma^2$ is not used in the loss function in Equation~\eqref{eq:mse}.
To estimate the variance, we can again maximize the likelihood of Equation~\eqref{eq:gaussian_likelihood} and obtain the maximum likelihood estimation of $\sigma^2$:
$
  \widehat{\sigma^2}_{ML} 
= 
  L(\widehat{W}_{ML})
$.

\subsection{Gaussian distribution}

Gaussian distributions have desired properties, including
\begin{enumerate}[itemsep=-1mm, topsep=-1mm, label=(\Alph*)]
\item 
\emph{stability}:
the family of Gaussian distributions is closed under affine transformations:
\begin{equation}
  a Z + b 
\sim 
  \Normal(a\mu + b, a^2\sigma^2)
;
\end{equation}

\item 
\emph{decomposability}:
the sum of independent Gaussian random variables is still Gaussian:
\begin{equation}
  Z_1 + Z_2 
\sim 
  \Normal(\mu_1 + \mu_2, 
          \sigma_1^2 + \sigma_2^2)
.
\end{equation}

\end{enumerate}

Based on these two properties, any linear combination of independent Gaussian random variables is still Gaussian distributed.
Thus, several aggregate observations such as the mean observation and the rank observation can be easily modeled and their log-likelihood has an analytical expression, as discussed in Section~\ref{ssec:method_mean} and Section~\ref{ssec:method_rank}.

%%%%%%%%%%%%%%%%%%%%%%%%%%%%%%%%%%%%%%%%%%%%%%%%%%

\section{Mean observation}
\label{app:mean}

In Section~\ref{ssec:method_mean}, we discussed a probabilistic model based on the Gaussian distribution for regression via mean observation.
In this part, we show that it is possible to use other probabilistic models such as Cauchy distribution and Poisson distribution for regression.

Here, $f_Z$ and $F_Z$ denote the probability density function (PDF) and the cumulative distribution function (CDF) of a real-valued random variable $Z$, respectively.

%%%%%%%%%%%%%%%%%%%%%%%%%%%%%%%%%%%%%%%%%%%%%%%%%%

\subsection{Cauchy distribution}
\label{sapp:cauchy}

Similarly to the Gaussian distribution, any distribution in the stable distribution family has the property that the linear combination of two independent random variables follows the same distribution up to location and scale parameters.
Among the stable distribution family, the Cauchy distribution is studied for the robust regression problem in \citet{liu2014robustness}, called Cauchy regression.

Concretely, 
let $Z \sim \Cauchy(a, b) \in \R$
with location parameter $a \in \R$ 
and scale parameter $b \in (0, \infty)$,
where
\begin{equation}
  f_{Z}(z) = \frac{1}{\pi b(1 + z'^2)}
,
\text{ where } 
  z' = \frac{z - a}{b}
,
\end{equation}
\begin{equation}
  F_{Z}(z) = \frac{1}{\pi} \arctan(z') + \frac12
.
\end{equation}
Then the mean 
$Y = T_{\mathrm{mean}}(Z_{1:K}) = \frac{1}{K} \sum_i^K Z_i$ 
is still Cauchy distributed:
\begin{equation}
  Y \sim 
  \Cauchy(\frac{1}{K} \sum_i^K a_i, \frac{1}{K} \sum_i^K b_i)
.
\end{equation}

%%%%%%%%%%%%%%%%%%%%%%%%%%%%%%%%%%%%%%%%%%%%%%%%%%

\subsection{Poisson distribution}

In some problems, the true target and the aggregate observations are non-negative integers, e.g., count data. 
It is different from the normal regression since the support space $\sZ$ is discrete.

Poisson distribution is popular for modeling such data, which is used in \citet{law2018variational} for spatially aggregated data.
Concretely, 
let $Z \sim \Poisson(\lambda) \in \N_0$ 
with rate parameter $\lambda \in (0, \infty)$, 
where
\begin{equation}
  f_{Z}(z) = \frac{\lambda^z e^{-\lambda}}{z!}
.
\end{equation}
Then the sum 
$Y = T_{\mathrm{sum}}(Z_{1:K}) = \sum_{i=1}^K Z_i$
is still Poisson distributed:
\begin{equation}
  Y \sim 
  \Poisson(\sum_{i=1}^K \lambda_i)
.
\end{equation}

%%%%%%%%%%%%%%%%%%%%%%%%%%%%%%%%%%%%%%%%%%%%%%%%%%

\section{Rank observation}
\label{app:rank}

%%%%%%%%%%%%%%%%%%%%%%%%%%%%%%%%%%%%%%%%%%%%%%%%%%

\subsection{Related tasks}

Regression via rank observation problem discussed in Section~\ref{ssec:method_rank} is different from learning to rank problem in terms of the type of training data and test data.
The exact value of the latent score in learning to rank is not important as long as the relative order is preserved,
but the output of a regression function is compared to the ground-truth so that not only the order but also the scale need to be predicted.
Another related task is \emph{ordinal regression} \citep{gutierrez2015ordinal}, where the target is usually discrete and ordered, but only the relative order between different values is important.
The type of training and test data of several supervised learning tasks is listed in Table~\ref{tab:tasks}, and examples of the type of measurements are given in Table~\ref{tab:measurements}.

\begin{table}[ht]
\caption{\textbf{The type of training and test data of supervised learning tasks}}
\label{tab:tasks}
\begin{center}
\begin{tabular}{lll}
\toprule
Task & Training data & Test data
\\ 
\midrule
Classification & categorical & categorical 
\\
Ordinal Regression & ordinal & ordinal 
\\
Regression & continuous & continuous 
\\
Regression via Rank & rank & continuous 
\\
Learning to Rank & continuous/rank & rank
\\
\bottomrule
\end{tabular}
\end{center}
\end{table}

\begin{table}[ht]
\caption{\textbf{Examples of the type of measurements}}
\label{tab:measurements}
\begin{center}
\begin{tabular}{llll}
\toprule
Measurement & Properties & Operators & Example
\\
\midrule
categorical 
& discrete, unordered
& $=$, $\neq$
& dog, cat, bear, otter, \dots
\\
\begin{tabular}[c]{@{}l@{}}
ordinal\\(rank)
\end{tabular}
& \begin{tabular}[c]{@{}l@{}}
discrete, ordered\\(binary)
\end{tabular}
& $>$, $<$
& 
\begin{tabular}[c]{@{}l@{}}
excellent, great, good, fair, bad\\(high, low)
\end{tabular}
\\
continuous
& continuous
& $+$, $-$
& $-2.71$, $0$, $3.14$, $\dots$
\\
\bottomrule
\end{tabular}
\end{center}
\end{table}

%%%%%%%%%%%%%%%%%%%%%%%%%%%%%%%%%%%%%%%%%%%%%%%%%%

\subsection{Gumbel distribution}

In Section~\ref{ssec:method_rank}, we discussed a probabilistic model based on the Gaussian distribution for regression via rank observation.
In this part, we provide other possible probabilistic models where we can also obtain an analytical expression of the log-likelihood.

Gumbel distributed latent scores are used implicitly in RankNet \citep{burges2005learning} and ListNet \citep{cao2007learning}.
Concretely, let 
$ Z \sim \Gumbel(\alpha, \beta) \in \R $
with location parameter $ \alpha \in \R $
and scale parameter $ \beta \in (0, \infty) $.
Its PDF and CDF are defined by
\begin{equation}
  f_{Z}(z) = \frac{1}{\beta} e^{-(z' + e^{-z'})}
,
\text{ where } 
  z' = \frac{z - \alpha}{\beta}
,
\end{equation}
\begin{equation}
  F_{Z}(z) = e^{-e^{-z'}}
.
\end{equation}
Then,
if two Gumbel distributed random variable $Z_1$ and $Z_2$ have the same scale $\beta$, we can derive that the difference of two Gumbel distributed random variables follows a logistic distribution:
\begin{equation}
  Z_1 - Z_2
\sim
  \Logistic(m = \alpha_1 - \alpha_2, b = \beta)
,
\end{equation}
where
\begin{equation}
  f_{Z_1 - Z_2}(d) = \frac{e^{-d'}}{\beta(1 + e^{-d'})^2}
,
\text{ where } 
  d' = \frac{d - m}{b}
,
\end{equation}
\begin{equation}
  F_{Z_1 - Z_2}(d) = \frac{1}{1 + e^{-d'}} 
.
\end{equation}

Similarly to Equation~\eqref{eq:rank}, we can derive that
\begin{equation}
  p(Z_1 > Z_2)
= 1 - F_{Z_1 - Z_2}(0)
= \frac{1}{1 + e^{-\frac{1}{\beta}(\alpha_1 - \alpha_2)}}
.
\end{equation}

Let $ \alpha = s $ be the output of the function $f(X; W)$ and fix the scale $ \beta = 1 $, we can get the probability of the rank
\begin{equation}
  p(Z_1 > Z_2) 
= \frac{1}{1 + e^{-(s_1 - s_2)}}
= \frac{e^{s_1}}{e^{s_1} + e^{s_2}}
.
\end{equation}
The negative log-likelihood is used as the loss function of RankNet \citep{burges2005learning}.

Note that the probability is in the form of the logistic function.
The scales $\beta$ should be the same for different Gumbel random variables in order to have an analytical expression of the likelihood of the difference.

%%%%%%%%%%%%%%%%%%%%%%%%%%%%%%%%%%%%%%%%%%%%%%%%%%

In some applications, rank observations are in the form of the permutation of a list instead of pairwise comparisons, e.g., 
$ Z_1 > Z_2 > \dots > Z_K $.
A naive way to utilize such data is to decompose it to pairwise comparisons.
However, the size of all possible pairs is
$ \binom{K}{2} = \frac{K(K-1)}{2} $,
which could be too large if $K$ is large.

On the other hand, for the Gumbel random variables, we could simplify the calculation based on some properties of the Gumbel distribution, including
\begin{enumerate}[itemsep=-1mm, topsep=-1mm, label=(\Alph*)]
\item 
\emph{max-stability}:
the maximum of a set of independent Gumbel distributed random variables is still Gumbel distributed:

Let
$
  \overline{Z} 
= 
  \max\{Z_1, \dots, Z_K\}
$,
then
\begin{equation}
  \overline{Z} 
\sim 
  \Gumbel(\overline{\alpha}, \beta)
,
\text{ where } 
  \overline{\alpha} 
= 
  \beta \log \sum_{i=1}^K \exp{\frac{\alpha_i}{\beta}}
.
\end{equation}

Therefore,
\begin{equation}
\label{eq:lca}
  p(Z_1 > \max\{Z_{2:K}\})
=
  \frac{1}{1 + \sum_{i=2}^K e^{-(s_1 - s_i)}}
= 
  \frac{e^{s_1}}{\sum_{i=1}^K e^{s_i}}
;
\end{equation}

\item
$
  p(Z_1 > \max\{Z_2, Z_3\})
=
  p(Z_1 > Z_2 | Z_2 > Z_3)
=
  p(Z_1 > Z_3 | Z_3 > Z_2)
$:

\begin{proof}

\begin{equation}
  p(Z_1 > \max\{Z_2, Z_3\})
=
  \frac{1}
  {1 
  + e^{-\frac1\beta \parens*{
  \alpha_1 - \beta\log
  \parens*{e^\frac{\alpha_2}{\beta} 
          +e^\frac{\alpha_3}{\beta}}
  }
  }}
=
  \frac
  {e^\frac{\alpha_1}{\beta}}
  {e^\frac{\alpha_1}{\beta} 
  +e^\frac{\alpha_2}{\beta} 
  +e^\frac{\alpha_3}{\beta}}
.  
\end{equation}

\begin{equation}
\begin{aligned}
  p(Z_1 > Z_2 > Z_3)
&=
  \intinf 
  \int_{-\infty}^{z_1} 
  \int_{-\infty}^{z_2}
  f_{Z_1}(z_1) f_{Z_2}(z_2) f_{Z_3}(z_3)
  \D{z_3} \D{z_2} \D{z_1}
\\
&=
  \frac
  {e^\frac{\alpha_1}{\beta}}
  {e^\frac{\alpha_1}{\beta} 
  +e^\frac{\alpha_2}{\beta} 
  +e^\frac{\alpha_3}{\beta}}
  \frac
  {e^\frac{\alpha_2}{\beta}}
  {e^\frac{\alpha_2}{\beta} 
  +e^\frac{\alpha_3}{\beta}}
\\
&=
  p(Z_1 > \max\{Z_2, Z_3\}) p(Z_2 > Z_3)
.
\end{aligned}
\end{equation}

Therefore
$
  p(Z_1 > Z_2 | Z_2 > Z_3)
=
  \frac{p(Z_1 > Z_2 > Z_3)}{p(Z_2 > Z_3)}
=
  p(Z_1 > \max\{Z_2, Z_3\})
$.

We can prove the second equation in a similar way.

\end{proof}

\end{enumerate}

Based on these two properties, it is trivial to generalize it and prove that for Gumbel random variables following property holds:
\begin{equation}
  p(Z_1 > Z_2 > \dots > Z_K)
=
  \prod_{i=1}^{K-1}
  p(Z_i > \max\{Z_{i+1:K}\})
,
\end{equation}
which is used in the loss function of ListNet \citep{cao2007learning}.
However, other distributions such as the Gaussian distribution and the Cauchy distribution do not have these properties, so it is hard to extend pairwise rank observation to listwise rank observation in this way for these distributions.

Note that some special cases of our framework are studied in psychometrics and econometrics, including
Bradley-Terry model (pairwise, Gumbel) \citep{bradley1952rank},
Plackett-Luce model (listwise, Gumbel)
\citep{plackett1975analysis},
and Thurstone-Mosteller model (pairwise, Gaussian) 
\citep{thurstone1927law},
which inspired algorithms such as RankNet \citep{burges2005learning} and ListNet \citep{cao2007learning}.
However, the focus is not on the underlying probabilistic model, e.g., the property shown in Equation~\ref{eq:lca} is treated as an axiom called Luce's choice axiom (LCA) \citep{luce1959individual} in economics rather than the property of the Gumbel distribution.
Our work clarifies the assumptions and the underlying probabilistic model and extends existing work to a variety of aggregate observations.

%%%%%%%%%%%%%%%%%%%%%%%%%%%%%%%%%%%%%%%%%%%%%%%%%%

\subsection{Exponential distribution}

In addition to the interpretation of Gumbel distributed latent scores, we provide another probabilistic model based on the exponential distribution that leads to the same learning objective.

Let $ Z \sim \Exponential(\lambda) \in [0, \infty) $
with inverse scale parameter $ \lambda \in (0, \infty) $,
where
\begin{equation}
  f_{Z}(z) = \lambda e^{-\lambda z}
,
\end{equation}
\begin{equation}
  F_{Z}(z) = 1 - e^{-\lambda z}
.
\end{equation}

Then, we can derive that the difference of two exponential distributed random variables follows a asymmetric Laplace distribution:
\begin{equation}
  Z_1 - Z_2
\sim
  \Asymmetric\Laplace\parens*{
  m = 0, 
  \lambda = \sqrt{\lambda_1 \lambda_2},
  \kappa = \sqrtfrac{\lambda_1}{\lambda_2}
  }
,
\end{equation}
where
\begin{equation}
  f_{Z_1 - Z_2}(d) 
= 
  \frac{\lambda_1 \lambda_2}{\lambda_1 + \lambda_2} 
  \begin{cases}
  e^{\lambda_2 d} & d < 0
  \\
  e^{-\lambda_1 d} & d \geq 0
  \end{cases}
,
\end{equation}
\begin{equation}
  F_{Z_1 - Z_2}(d)
=
  \begin{cases}
  \frac{\lambda_1}{\lambda_1 + \lambda_2} 
  e^{\lambda_2 d} & d < 0
  \\
  1 - \frac{\lambda_2}{\lambda_1 + \lambda_2} 
  e^{-\lambda_1 d} & d \geq 0
  \end{cases}
.
\end{equation}

Similarly to Equation~\ref{eq:rank}, we can obtain the probability of the rank
\begin{equation}
  p(Z_1 > Z_2)
=
  1 - F_{Z_1 - Z_2}(0)
=
  \frac{\lambda_2}{\lambda_1 + \lambda_2}
.
\end{equation}

Let $ \lambda = e^{-s} $ and let $s$ be the output of the function $f(X; W)$,
then
\begin{equation}
  p(Z_1 > Z_2) 
= 
  \frac{1}{1 + e^{-(s_1 - s_2)}}
=
  \frac{e^{s_1}}{e^{s_1} + e^{s_2}}
.
\end{equation}

Note that $ -\log(Z) \sim \Gumbel(-\log\lambda, 1) $, which relates to the interpretation of Gumbel distributed latent scores.

If we use other positive function instead of $ e^{-s} $ that maps the output of function $f(X; W)$ to the inverse scale parameter $\lambda$, we can derive different learning objectives.
The extension is left for future work.

%%%%%%%%%%%%%%%%%%%%%%%%%%%%%%%%%%%%%%%%%%%%%%%%%%

\subsection{Cauchy distribution}

As discussed in \ref{sapp:cauchy}, Cauchy can be used for robust regression.
It also can be used for regression via rank, because the difference of two Cauchy random variables is again Cauchy distributed:
\begin{equation}
  Z_1 - Z_2
\sim
  \Cauchy\parens*{
  a = a_1 - a_2,
  b = b_1 + b_2
  }
,
\end{equation}
while the probability of the rank is
\begin{equation}
  p(Z_1 > Z_2)
=
  1 - F_{Z_1 - Z_2}(0)
=
  \frac{1}{\pi} 
  \arctan\parens*{\frac{a_1 - a_2}{b_1 + b_2}} 
+ \frac12
.
\end{equation}

In conclusion, different distribution assumptions correspond to different sigmoid functions in regression via rank observation and learning to rank problems.
Gumbel distributions correspond to the logistic function (the hyperbolic tangent), 
Gaussian distributions correspond to the error function,
and Cauchy distributions correspond to the arctangent function.

For regression via rank observation, the distribution assumption should match the data distribution, especially when we assume that the noise is homoscedastic so the variance is a fixed value.
For learning to rank, the Gumbel distribution allows for modeling the maximum of a set of scores explicitly.
On the other hand, it is more flexible to model the uncertainty of scores using two-parameter Gaussian distribution or Cauchy distribution if we assume that the noise is heteroscedastic.
Because of the fat tails of the Cauchy distribution, it allows for values far from the expected value so that learning could be more robust.

%%%%%%%%%%%%%%%%%%%%%%%%%%%%%%%%%%%%%%%%%%%%%%%%%%

\clearpage
\section{Experiment details}
\label{app:experiment_details}

In this section, we provide missing experiment details in Section~\ref{sec:experiments}.

%%%%%%%%%%%%%%%%%%%%%%%%%%%%%%%%%%%%%%%%%%%%%%%%%%

\subsection{Classification via triplet comparison (Section~\ref{ssec:experiment_triplet})}

%%%%%%%%%%%%%%%%%%%%
\paragraph{Data.}
We used MNIST \citep{lecun1998gradient}, Fashion-MNIST \citep{xiao2017fashion} and Kuzushiji-MNIST \citep{clanuwat2018deep} datasets without any data augmentation.
They all contain $28 \times 28$ grayscale images in $10$ classes.
The size of the original training dataset is $60000$ and the size of the original test dataset is $10000$.

%%%%%%%%%%%%%%%%%%%%
\paragraph{Data preprocessing.}
For both pairwise similarity and triplet comparison data, pairs and triplets of data were sampled randomly with replacement from the original training dataset according to our assumptions in Section~\ref{ssec:assumptions}.
The size of generated datasets for direct, pairwise, and triplet observations are $60000$, $120000$, and $180000$, respectively.

%%%%%%%%%%%%%%%%%%%%
\paragraph{Model.}
We used a sequential convolutional neural network as the model $f(X; W)$:
Conv2d(\#channel = 32),
ReLU,
Conv2d(\#channel = 64),
MaxPool2d(size = 2),
Dropout(p = 0.25),
Linear(\#dim = 128),
ReLU,
Dropout(p = 0.5),
Linear(\#dim = 10).
The kernel size of convolutional layers is $3$,
and the kernel size of max pooling layer is $2$.

%%%%%%%%%%%%%%%%%%%%
\paragraph{Optimization.}
We used the Adam optimizer with decoupled weight decay regularization \citep{loshchilov2018decoupled} to train the model.
The learning rate is \num{1e-3},
the batch size is $128$,
and the model is trained for $10$ epochs.

%%%%%%%%%%%%%%%%%%%%%%%%%%%%%%%%%%%%%%%%%%%%%%%%%%

\subsection{Regression via mean/rank observation on UCI datasets (Section~\ref{ssec:experiment_mean_rank})}

%%%%%%%%%%%%%%%%%%%%
\paragraph{Data.}
We used $6$ benchmark regression datasets from the UCI machine learning repository \citep{UCI}.
Table~\ref{tab:uci_info} shows the statistic of thsese datasets. 

\begin{table}[ht]
\centering
\caption{\textbf{Statistic of 6 UCI Benchmark Datasets.}}
\label{tab:uci_info}
\begin{tabular}{ccccc}
\toprule
Dataset & Dimension & Number of data \\ 
\midrule
abalone & 10 & 353 
\\
airfoil & 5 & 1202
\\
auto-mpg & 7 & 313 
\\
concrete & 8 & 824
\\
housing & 13 & 404
\\
power-plant & 4 & 7654
\\
\bottomrule
\end{tabular}
\end{table}

%%%%%%%%%%%%%%%%%%%%
\paragraph{Data preprocessing.}
We split the original datasets into training, validation, and test sets randomly by $60\%$, $20\%$, and $20\%$ for each trial. 
Only training sets are used for generating the mean and the rank observations.
The feature vectors $X$ are standardized to have $0$ mean and $1$ standard deviation and the targets $Z$ are normalized to have $0$ mean using statistics of training sets.
For mean observations, the number of sets is the same as the the number of original datasets and each set contains four instances.
For rank observations, ten times of data are generated.

%%%%%%%%%%%%%%%%%%%%
\paragraph{Model.}
We used linear model and gradient boosting machine as the model $f(X; W)$.
We used LightGBM \citep{ke2017lightgbm} to implement the gradient boosting machine as the model $f(X; W)$.

%%%%%%%%%%%%%%%%%%%%
\paragraph{Optimization.}
For the linear model, we used stochastic gradient descent (SGD) with no momentum to train the model.
The learning rate is $0.1$,
the batch size is $256$,
and the model is trained for $20$ epochs.
For the gradient boosting machines, 
the boosting type is gradient boosting decision trees (GBDT),
the number of boost rounds is $100$,
and the number of early stopping rounds is $20$.

%%%%%%%%%%%%%%%%%%%%%%%%%%%%%%%%%%%%%%%%%%%%%%%%%%

\clearpage
\section{Additional experiments}
\label{app:additional_experiments}

In this section, we provide additional experimental results on the datasets different from Section~\ref{sec:experiments}.
The experiment setup is identical to that of the experiment in Section~\ref{sec:experiments}.
For the classification task, we used 20 binary classification datasets and 10 multiclass classification datasets. 
For the regression task, we used additional 20 regression datasets. 

%%%%%%%%%%%%%%%%%%%%%%%%%%%%%%%%%%%%%%%%%%%%%%%%%%

\subsection{Classification}
We conduct additional experiments on classification via pairwise similarity and triplet comparison in both binary and multiclass classification settings.

%%%%%%%%%%%%%%%%%%%%%%%%%%%%%%%%%%%%%%%%%%%%%%%%%%

\subsubsection{Binary classification}
Table~\ref{tab:additional_classification_binary_info} shows the statistic of the additional binary classification datasets. 
The datasets are obtained from OpenML~\citep{OpenML2013}, the UCI machine learning repository~\citep{UCI} and LIBSVM~\citep{libsvm}. 
Table~\ref{tab:additional_classification_binary} shows the experimental results. 
For the pairwise similarity case, we can see that contrastive learning and our method can perform comparably to the supervised method in many datasets when we have only two classes.
For the triplet comparison, our method clearly outperformed other methods in most cases. 
Interestingly, Tuplet baseline performs relatively well and arguably better than the Triplet baseline in the binary case. 
As we will see in the multiclass case, the performance of Tuplet drops miserably compared with other methods.

\begin{table}[ht]
\caption{\textbf{Statistic of 20 Additional Binary Classification Datasets.}}
\label{tab:additional_classification_binary_info}
\centering
\begin{tabular}{ccccc}
\toprule
Dataset & Dimension & Positive & Negative & Number of data \\ 
\midrule
adult & 104 & 7508 & 22652 & 30160 \\
ayi & 100 & 1385 & 1362 & 2747 \\
banana & 2 & 2376 & 2923 & 5299 \\
codrna & 8 & 162855 & 325709 & 488564 \\
custrev & 100 & 2405 & 1365 & 3770 \\
ijcnn & 22 & 18418 & 173262 & 191680 \\
image & 18 & 1187 & 898 & 2085 \\
magic & 10 & 12330 & 6687 & 19017 \\
mpqa & 100 & 3311 & 7291 & 10602 \\
mushroom & 98 & 3487 & 2155 & 5642 \\
phishing & 30 & 6157 & 4896 & 11053 \\
phoneme & 5 & 3817 & 1586 & 5403 \\
ringnorm & 20 & 3663 & 3736 & 7399 \\
rt-polarity & 100 & 5331 & 5330 & 10661 \\
spambase & 57 & 1811 & 2788 & 4599 \\
splice & 60 & 1344 & 1646 & 2990 \\
subj & 100 & 5000 & 4999 & 9999 \\
susy & 18 & 45974 & 54024 & 99998 \\
twonorm & 20 & 3703 & 3696 & 7399 \\
w8a & 300 & 1933 & 62766 & 64699 \\
\bottomrule
\end{tabular}
\end{table}

\begin{table}[ht]
\caption{
  \textbf{Classification via pairwise similarity and triplet comparison.}
  Means and standard deviations of accuracy in percentage for $10$ trials are reported.
}
\label{tab:additional_classification_binary}
\centering
\resizebox{\textwidth}{!}{%
\begin{tabular}{lcccclcccc}
\toprule
Dataset & 
\multicolumn{1}{c}{Unsupervised} & \multicolumn{3}{c}{Pairwise Similarity} & 
& 
\multicolumn{3}{c}{Triplet Comparison} & \multicolumn{1}{c}{Supervised} 
\\ 
\cmidrule{3-5} 
\cmidrule{7-9} 
& & 
Siamese & Contrastive & Ours & 
& 
Tuplet & Triplet & Ours &  
\\ 
\midrule
adult
& $71.38$ &$69.91$ &$81.86$ &$\mathbf{84.15}$ && $73.73$ &$57.89$ &$\mathbf{75.49}$ &$85.08$ \\
& $(0.29)$ &$(11.11)$ &$(3.57)$ &$(\mathbf{0.43})$ && $(0.59)$ &$(7.07)$ &$(\mathbf{0.63})$ &$(0.29)$ \\
ayi
& $53.25$ &$66.93$ &$\mathbf{76.67}$ &$\mathbf{77.65}$ && $52.16$ &$\mathbf{54.22}$ &$\mathbf{54.51}$ &$78.47$ \\
& $(1.58)$ &$(10.90)$ &$(\mathbf{1.20})$ &$(\mathbf{1.68})$ && $(2.09)$ &$(\mathbf{1.40})$ &$(\mathbf{3.28})$ &$(1.40)$ \\
banana
& $56.16$ &$70.22$ &$\mathbf{89.80}$ &$\mathbf{89.47}$ && $53.26$ &$\mathbf{56.75}$ &$\mathbf{57.53}$ &$88.42$ \\
& $(0.94)$ &$(16.99)$ &$(\mathbf{1.20})$ &$(\mathbf{0.79})$ && $(2.50)$ &$(\mathbf{6.25})$ &$(\mathbf{5.36})$ &$(1.11)$ \\
codrna
& $55.58$ &$59.23$ &$96.38$ &$\mathbf{96.57}$ && $61.06$ &$59.28$ &$\mathbf{66.72}$ &$96.47$ \\
& $(0.10)$ &$(12.33)$ &$(0.08)$ &$(\mathbf{0.11})$ && $(2.43)$ &$(8.68)$ &$(\mathbf{0.18})$ &$(0.07)$ \\
custrev
& $51.70$ &$60.76$ &$70.84$ &$\mathbf{74.56}$ && $53.24$ &$52.33$ &$\mathbf{60.92}$ &$75.32$ \\
& $(1.26)$ &$(9.07)$ &$(1.27)$ &$(\mathbf{1.10})$ && $(1.67)$ &$(2.13)$ &$(\mathbf{4.67})$ &$(2.13)$ \\
ijcnn
& $58.39$ &$58.48$ &$\mathbf{99.13}$ &$98.98$ && $74.02$ &$52.88$ &$\mathbf{92.15}$ &$99.08$ \\
& $(1.65)$ &$(13.02)$ &$(\mathbf{0.04})$ &$(0.19)$ && $(10.74)$ &$(4.34)$ &$(\mathbf{0.56})$ &$(0.05)$ \\
image
& $60.94$ &$75.44$ &$\mathbf{94.10}$ &$\mathbf{93.81}$ && $54.51$ &$\mathbf{56.09}$ &$\mathbf{59.02}$ &$89.76$ \\
& $(1.44)$ &$(18.45)$ &$(\mathbf{0.84})$ &$(\mathbf{1.16})$ && $(3.15)$ &$(\mathbf{3.31})$ &$(\mathbf{7.21})$ &$(0.99)$ \\
magic
& $54.55$ &$77.18$ &$86.44$ &$\mathbf{86.93}$ && $\mathbf{67.49}$ &$58.79$ &$65.94$ &$86.31$ \\
& $(0.81)$ &$(12.22)$ &$(0.59)$ &$(\mathbf{0.52})$ && $(\mathbf{2.50})$ &$(5.27)$ &$(1.08)$ &$(0.41)$ \\
mpqa
& $62.56$ &$59.38$ &$\mathbf{84.76}$ &$\mathbf{84.60}$ && $53.33$ &$53.18$ &$\mathbf{73.06}$ &$85.89$ \\
& $(0.86)$ &$(13.56)$ &$(\mathbf{0.68})$ &$(\mathbf{0.87})$ && $(2.73)$ &$(1.24)$ &$(\mathbf{1.41})$ &$(0.67)$ \\
mushroom
& $85.14$ &$82.17$ &$\mathbf{99.98}$ &$\mathbf{99.96}$ && $61.37$ &$56.47$ &$\mathbf{88.46}$ &$99.97$ \\
& $(0.75)$ &$(21.95)$ &$(\mathbf{0.04})$ &$(\mathbf{0.08})$ && $(4.28)$ &$(6.04)$ &$(\mathbf{8.43})$ &$(0.08)$ \\
phishing
& $54.80$ &$77.65$ &$\mathbf{96.07}$ &$\mathbf{96.43}$ && $55.64$ &$57.08$ &$\mathbf{73.28}$ &$95.25$ \\
& $(0.82)$ &$(21.31)$ &$(\mathbf{0.37})$ &$(\mathbf{0.56})$ && $(3.02)$ &$(4.82)$ &$(\mathbf{2.44})$ &$(0.35)$ \\
phoneme
& $68.24$ &$60.64$ &$80.82$ &$\mathbf{83.59}$ && $59.16$ &$62.16$ &$\mathbf{71.39}$ &$81.06$ \\
& $(0.51)$ &$(9.46)$ &$(1.00)$ &$(\mathbf{1.27})$ && $(8.19)$ &$(8.20)$ &$(\mathbf{0.72})$ &$(0.75)$ \\
ringnorm
& $76.05$ &$82.95$ &$97.20$ &$\mathbf{98.07}$ && $\mathbf{72.54}$ &$58.82$ &$\mathbf{67.92}$ &$97.97$ \\
& $(1.05)$ &$(20.56)$ &$(0.39)$ &$(\mathbf{0.22})$ && $(\mathbf{4.11})$ &$(8.26)$ &$(\mathbf{9.70})$ &$(0.31)$ \\
rt-polarity
& $51.99$ &$57.44$ &$69.73$ &$\mathbf{71.04}$ && $52.26$ &$\mathbf{52.82}$ &$\mathbf{54.20}$ &$72.37$ \\
& $(0.82)$ &$(8.92)$ &$(0.71)$ &$(\mathbf{0.53})$ && $(1.64)$ &$(\mathbf{2.27})$ &$(\mathbf{2.53})$ &$(1.01)$ \\
spambase
& $60.04$ &$81.70$ &$93.61$ &$\mathbf{94.23}$ && $57.66$ &$60.00$ &$\mathbf{71.85}$ &$93.73$ \\
& $(1.17)$ &$(18.05)$ &$(0.44)$ &$(\mathbf{0.48})$ && $(1.50)$ &$(6.13)$ &$(\mathbf{7.39})$ &$(0.88)$ \\
splice
& $66.69$ &$70.15$ &$87.79$ &$\mathbf{90.95}$ && $53.16$ &$54.23$ &$\mathbf{59.05}$ &$89.48$ \\
& $(1.25)$ &$(19.07)$ &$(1.16)$ &$(\mathbf{1.23})$ && $(1.59)$ &$(2.78)$ &$(\mathbf{7.63})$ &$(1.11)$ \\
subj
& $82.28$ &$74.99$ &$87.48$ &$\mathbf{88.79}$ && $53.12$ &$54.79$ &$\mathbf{64.36}$ &$89.18$ \\
& $(0.86)$ &$(16.71)$ &$(0.56)$ &$(\mathbf{0.53})$ && $(2.42)$ &$(3.60)$ &$(\mathbf{6.80})$ &$(0.85)$ \\
susy
& $67.19$ &$66.07$ &$79.48$ &$\mathbf{79.78}$ && $\mathbf{58.68}$ &$53.67$ &$54.65$ &$79.87$ \\
& $(0.27)$ &$(12.93)$ &$(0.37)$ &$(\mathbf{0.20})$ && $(\mathbf{2.84})$ &$(2.93)$ &$(4.60)$ &$(0.26)$ \\
twonorm
& $97.66$ &$84.36$ &$\mathbf{97.64}$ &$\mathbf{97.64}$ && $\mathbf{60.60}$ &$\mathbf{58.16}$ &$\mathbf{63.47}$ &$97.65$ \\
& $(0.36)$ &$(20.17)$ &$(\mathbf{0.37})$ &$(\mathbf{0.42})$ && $(\mathbf{8.76})$ &$(\mathbf{2.99})$ &$(\mathbf{9.13})$ &$(0.37)$ \\
w8a
& $96.38$ &$61.49$ &$\mathbf{98.00}$ &$\mathbf{98.26}$ && $96.97$ &$57.98$ &$\mathbf{97.78}$ &$99.05$ \\
& $(1.79)$ &$(12.33)$ &$(\mathbf{0.59})$ &$(\mathbf{1.00})$ && $(0.15)$ &$(2.36)$ &$(\mathbf{0.15})$ &$(0.04)$ \\
\bottomrule
\end{tabular}
} % \resizebox
\end{table}

%%%%%%%%%%%%%%%%%%%%%%%%%%%%%%%%%%%%%%%%%%%%%%%%%%

\clearpage
\subsubsection{Multiclass classification}
Table~\ref{tab:additional_classification_multiclass_info} shows the statistic of the additional multiclass classification datasets. The datasets are obtained from OpenML~\citep{OpenML2013}. 
In Table~\ref{tab:additional_classification_multiclass}, our methods nicely outperformed representation learning based methods in all cases for triplet comparison. 
For pairwise similarity, contrastive learning outperformed our method in cardiotocography and isolet datasets, while our method clearly outperformed other methods on artificial-character, covertypy, gas-drift, and satimage. 

\begin{table}[ht]
\centering
\caption{\textbf{Statistic of 10 Additional Multiclass Datasets.}}
\label{tab:additional_classification_multiclass_info}
\begin{tabular}{ccccc}
\toprule
Dataset & Dimension & Number of classes & Number of data \\ 
\midrule
artificial-character & 7 & 10 & 10217 &  \\
cardiotocography & 35 & 10 & 2125 &  \\
covertype & 54 & 7 & 581011 &  \\
gas-drift & 128 & 6 & 13909 &  \\
isolet & 617 & 26 & 7796 &  \\
japanesevowels & 14 & 9 & 9960 &  \\
letter & 16 & 26 & 19999 &  \\
pendigits & 16 & 10 & 10991 &  \\
satimage & 36 & 6 & 6429 &  \\
vehicle & 18 & 4 & 845 &  \\
\bottomrule
\end{tabular}
\end{table}

\begin{table}[ht]
\caption{
  \textbf{Classification via pairwise similarity and triplet comparison.}
  Means and standard deviations of accuracy in percentage for $10$ trials are reported.
}
\label{tab:additional_classification_multiclass}
\centering
\resizebox{\textwidth}{!}{%
\begin{tabular}{lcccclcccc}
\toprule
Dataset & 
\multicolumn{1}{c}{Unsupervised} & \multicolumn{3}{c}{Pairwise Similarity} & 
& 
\multicolumn{3}{c}{Triplet Comparison} & \multicolumn{1}{c}{Supervised} 
\\ 
\cmidrule{3-5} 
\cmidrule{7-9} 
& & 
Siamese & Contrastive & Ours & 
& 
Tuplet & Triplet & Ours &  
\\ 
\midrule
artificial-character
& $22.05$ &$34.93$ &$46.31$ &$\mathbf{50.29}$ && $22.18$ &$23.04$ &$\mathbf{47.71}$ &$57.71$ \\
& $(0.75)$ &$(2.76)$ &$(0.78)$ &$(\mathbf{1.19})$ && $(1.97)$ &$(1.90)$ &$(\mathbf{1.56})$ &$(1.00)$ \\
cardiotocography
& $88.14$ &$85.88$ &$\mathbf{99.88}$ &$95.39$ && $36.89$ &$37.27$ &$\mathbf{94.35}$ &$99.95$ \\
& $(6.57)$ &$(15.75)$ &$(\mathbf{0.24})$ &$(1.73)$ && $(2.90)$ &$(4.52)$ &$(\mathbf{6.76})$ &$(0.09)$ \\
covertype
& $36.24$ &$47.81$ &$54.17$ &$\mathbf{81.15}$ && $36.14$ &$25.17$ &$\mathbf{64.85}$ &$84.69$ \\
& $(3.97)$ &$(6.17)$ &$(4.03)$ &$(\mathbf{1.09})$ && $(3.80)$ &$(2.89)$ &$(\mathbf{7.37})$ &$(0.35)$ \\
gas-drift
& $37.29$ &$61.32$ &$90.40$ &$\mathbf{98.40}$ && $26.32$ &$38.71$ &$\mathbf{82.60}$ &$98.56$ \\
& $(0.78)$ &$(15.06)$ &$(7.57)$ &$(\mathbf{0.59})$ && $(1.58)$ &$(4.86)$ &$(\mathbf{5.55})$ &$(0.33)$ \\
isolet
& $55.69$ &$73.04$ &$\mathbf{86.24}$ &$78.79$ && $17.32$ &$22.38$ &$\mathbf{57.73}$ &$96.12$ \\
& $(2.38)$ &$(1.88)$ &$(\mathbf{2.27})$ &$(3.44)$ && $(1.02)$ &$(2.90)$ &$(\mathbf{3.14})$ &$(0.57)$ \\
japanesevowels
& $38.42$ &$73.29$ &$\mathbf{96.25}$ &$\mathbf{96.61}$ && $27.80$ &$31.69$ &$\mathbf{94.66}$ &$96.55$ \\
& $(1.00)$ &$(12.45)$ &$(\mathbf{0.37})$ &$(\mathbf{0.55})$ && $(2.68)$ &$(3.49)$ &$(\mathbf{0.75})$ &$(0.38)$ \\
letter
& $29.06$ &$48.43$ &$\mathbf{73.59}$ &$\mathbf{73.86}$ && $21.08$ &$23.12$ &$\mathbf{64.56}$ &$89.41$ \\
& $(1.74)$ &$(5.10)$ &$(\mathbf{1.65})$ &$(\mathbf{2.14})$ && $(1.46)$ &$(1.95)$ &$(\mathbf{1.57})$ &$(0.57)$ \\
pendigits
& $68.22$ &$83.17$ &$\mathbf{99.03}$ &$\mathbf{99.02}$ && $43.01$ &$41.01$ &$\mathbf{97.24}$ &$98.68$ \\
& $(2.74)$ &$(8.09)$ &$(\mathbf{0.18})$ &$(\mathbf{0.22})$ && $(3.68)$ &$(3.08)$ &$(\mathbf{2.12})$ &$(0.20)$ \\
satimage
& $64.32$ &$65.41$ &$83.52$ &$\mathbf{88.23}$ && $44.28$ &$47.30$ &$\mathbf{84.33}$ &$87.82$ \\
& $(6.01)$ &$(4.21)$ &$(1.87)$ &$(\mathbf{0.52})$ && $(4.03)$ &$(6.61)$ &$(\mathbf{0.84})$ &$(0.84)$ \\
vehicle
& $38.40$ &$46.21$ &$\mathbf{69.35}$ &$\mathbf{68.34}$ && $35.92$ &$41.60$ &$\mathbf{57.10}$ &$69.23$ \\
& $(1.70)$ &$(7.79)$ &$(\mathbf{5.35})$ &$(\mathbf{4.90})$ && $(3.39)$ &$(5.63)$ &$(\mathbf{7.28})$ &$(2.66)$ \\
\bottomrule
\end{tabular}
} % \resizebox
\end{table}

%%%%%%%%%%%%%%%%%%%%%%%%%%%%%%%%%%%%%%%%%%%%%%%%%%

\clearpage
\subsection{Regression}

\subsubsection{Varying size of sets in regression via mean observation}
Table~\ref{tab:varying_k} shows the experimental result when the number of sample points that are aggregated in each set (i.e., $K$) is changed, where $K \in \{2, 4, 8, 16\}$.
The result shows that larger sets lead to slightly worse performance, which illustrates the trade-off between the quality of labels and the costs/privacy conservation. 
However, the performance difference is still marginal.

\begin{table}[h]
\caption{
\textbf{Varying $K$ in regression via mean observation on UCI benchmark datasets.} 
Means and standard deviations of MSE for $10$ trials are reported.
We compare linear regression (LR) and gradient boosting machines (GBM) as the regression function.
}
\label{tab:varying_k}
\centering
\begin{tabular}{lcccclcccc}
\toprule
Dataset & \multicolumn{4}{c}{LR} & & \multicolumn{4}{c}{GBM}
\\ 
\cmidrule{2-5} \cmidrule{7-10}
& $2$ & $4$ & $8$ & $16$ & 
& $2$ & $4$ & $8$ & $16$
\\ 
\midrule
abalone
& $5.52$ & $5.33$ & $5.60$ & $5.89$ & 
& $4.68$ & $4.89$ & $4.70$ & $4.98$ 
\\
& $(0.4)$ & $(0.4)$ & $(0.5)$ & $(0.4)$ & 
& $(0.3)$ & $(0.4)$ & $(0.2)$ & $(0.3)$ 
\\
airfoil
& $23.08$ & $24.00$ & $25.78$ & $30.91$ & 
& $4.87$ & $4.97$ & $4.98$ & $4.98$ 
\\
& $(2.0)$ & $(2.2)$ & $(1.5)$ & $(1.7)$ & 
& $(0.5)$ & $(0.7)$ & $(0.7)$ & $(0.8)$ 
\\
auto-mpg
& $13.54$ & $12.67$ & $15.00$ & $21.52$ & 
& $10.03$ & $9.88$ & $10.46$ & $10.68$ 
\\
& $(2.1)$ & $(2.3)$ & $(3.3)$ & $(4.6)$ & 
& $(2.6)$ & $(1.6)$ & $(2.5)$ & $(2.0)$ 
\\
concrete
& $120.36$ & $121.74$ & $119.78$ & $149.48$ & 
& $31.98$ & $32.01$ & $32.55$ & $31.71$ 
\\
& $(12.9)$ & $(13.3)$ & $(8.9)$ & $(12.8)$ & 
& $(3.8)$ & $(3.7)$ & $(3.4)$ & $(4.4)$ 
\\
housing
& $27.82$ & $28.36$ & $20.90$ & $34.67$ & 
& $13.27$ & $17.97$ & $16.63$ & $15.10$ 
\\
& $(9.5)$ & $(8.1)$ & $(4.4)$ & $(6.8)$ & 
& $(2.8)$ & $(4.4)$ & $(4.4)$ & $(5.1)$ 
\\
power-plant
& $21.02$ & $21.13$ & $21.61$ & $24.12$ & 
& $13.17$ & $13.04$ & $14.03$ & $13.49$ 
\\
& $(1.0)$ & $(1.4)$ & $(1.0)$ & $(0.8)$ & 
& $(0.7)$ & $(0.7)$ & $(1.2)$ & $(0.9)$ 
\\
\bottomrule
\end{tabular}
\end{table}

\subsubsection{Mean/rank observation}
Table~\ref{tab:additional_regression_info} shows the statistic of the additional regression datasets. 
The datasets are obtained from OpenML~\citep{OpenML2013} and the UCI machine learning repository~\citep{UCI}. 
Table~\ref{tab:additional_regression} shows the mean squared error results for each method. 
For the regression via mean observation, our method based on linear regression outperformed the linear regression baseline consistently in all cases, similarly for the gradient boosting machine case. 
Sometimes they are comparable to the supervised method.
For the regression via rank observation, it is interesting to see that our method based on a Gaussian distribution failed in house-16h and house-8l datasets. 
We hypothesize that it is because the target distribution is heavy-tailed in these datasets.

\begin{table}[ht]
\centering
\caption{\textbf{Statistic of 20 Additional Regression Datasets.}}
\label{tab:additional_regression_info}
\begin{tabular}{ccccc}
\toprule
Dataset & Dimension & Min & Max & Number of data \\ 
\midrule
2d-planes & 10 & -12.6943 & 12.2026 & 40767 \\
bank-32nh & 32 & 0 & 82 & 8191 \\
bank-8fm & 8 & 0 & 80.2263 & 8191 \\
bike-sharing & 33 & 0.022 & 8.71 & 730 \\
cpu-act & 21 & 0 & 99 & 8191 \\
diabetes & 10 & 25 & 346 & 441 \\
elevator & 6 & -15 & 15.1 & 9516 \\
fried & 10 & -1.228 & 30.522 & 40767 \\
house-16h & 16 & 0 & 50 & 22783 \\
house-8l & 8 & 0 & 50 & 22783 \\
insurance-charge & 11 & 1.12 & 63.8 & 1336 \\
kin-8nm & 8 & 4.016538 & 145.8521 & 8191 \\
puma-8nh & 8 & -12.4153 & 11.87619 & 8191 \\
real-estate & 6 & 7.6 & 117.5 & 413 \\
rmftsa-ladata & 10 & 4.15 & 30.43 & 507 \\
space-ga100 & 6 & -305.704 & 10.00835 & 3106 \\
stock & 9 & 34 & 62 & 949 \\
wine-quality & 11 & 3 & 9 & 6496 \\
wine-red & 11 & 3 & 8 & 1598 \\
wine-white & 11 & 3 & 9 & 4897 \\
\bottomrule
\end{tabular}
\end{table}

\begin{table}[ht]
\caption{
  \textbf{Regression via mean observation and rank observation.} 
  Means and standard deviations of error variance (rank observations) or MSE (otherwise) for $10$ trials are reported.
}
\label{tab:additional_regression}
\centering
\resizebox{\textwidth}{!}{%
\begin{tabular}{lcccclcccclcc}
\toprule
Dataset & 
\multicolumn{4}{c}{Mean Observation} & 
& 
\multicolumn{4}{c}{Rank Observation} &
&
\multicolumn{2}{c}{Supervised} 
\\ 
\cmidrule{2-5} \cmidrule{7-10}
& 
\multicolumn{2}{c}{Baseline} & 
\multicolumn{2}{c}{Ours} &
& 
\multicolumn{2}{c}{RankNet, Gumbel} &
\multicolumn{2}{c}{Ours, Gaussian} & 
&
\\ 
\cmidrule(lr){2-3} 
\cmidrule(lr){4-5} 
\cmidrule(lr){7-8} 
\cmidrule(lr){9-10} 
\cmidrule(lr){12-13}
 & LR & GBM & LR & GBM & 
 & LR & GBM & LR & GBM & 
 & LR & GBM 
\\ 
\midrule
2d-planes
& $13.49$ &$11.40$ &$5.70$ &$\mathbf{1.01}$ && $8.86$ &$19.14$ &$5.72$ &$\mathbf{1.03}$ && $5.72$ & $1.00$
\\
& $(0.20)$ &$(0.20)$ &$(0.00)$ &$(\mathbf{0.00})$ && $(0.10)$ &$(0.20)$ &$(0.10)$ &$(\mathbf{0.00})$ && $(0.1)$ & $(0.0)$
\\
bank-32nh
& $112.42$ &$111.55$ &$\mathbf{69.30}$ &$\mathbf{70.32}$ && $123.63$ &$142.97$ &$\mathbf{71.25}$ &$87.79$ && $70.62$ & $68.83$
\\
& $(5.70)$ &$(4.00)$ &$(\mathbf{2.60})$ &$(\mathbf{3.70})$ && $(6.50)$ &$(6.40)$ &$(\mathbf{3.90})$ &$(4.10)$ && $(3.1)$ & $(4.2)$
\\
bank-8fm
& $134.99$ &$133.56$ &$15.23$ &$\mathbf{10.07}$ && $86.92$ &$230.19$ &$\mathbf{34.19}$ &$\mathbf{33.17}$ && $15.45$ & $9.21$
\\
& $(5.70)$ &$(4.50)$ &$(0.90)$ &$(\mathbf{0.30})$ && $(2.60)$ &$(6.40)$ &$(\mathbf{1.20})$ &$(\mathbf{1.50})$ && $(0.7)$ & $(0.5)$
\\
bike-sharing
& $2.36$ &$2.40$ &$0.77$ &$\mathbf{0.63}$ && $1.61$ &$3.79$ &$0.68$ &$\mathbf{0.57}$ && $0.64$ & $0.45$
\\
& $(0.20)$ &$(0.20)$ &$(0.10)$ &$(\mathbf{0.10})$ && $(0.30)$ &$(0.30)$ &$(0.10)$ &$(\mathbf{0.10})$ && $(0.1)$ & $(0.1)$
\\
cpu-act
& $235.22$ &$200.64$ &$98.20$ &$\mathbf{6.10}$ && $231.24$ &$343.39$ &$26484.33$ &$\mathbf{145.88}$ && $102.68$ & $5.07$
\\
& $(23.70)$ &$(18.40)$ &$(10.20)$ &$(\mathbf{0.40})$ && $(10.60)$ &$(33.40)$ &$(18656.30)$ &$(\mathbf{12.70})$ && $(17.1)$ & $(0.2)$
\\
diabetes
& $4688.70$ &$4937.20$ &$\mathbf{3335.22}$ &$\mathbf{3746.14}$ && $5765.28$ &$5918.11$ &$5968.57$ &$\mathbf{3771.73}$ && $3323.98$ & $3462.89$
\\
& $(541.30)$ &$(687.70)$ &$(\mathbf{430.40})$ &$(\mathbf{782.30})$ && $(303.80)$ &$(700.90)$ &$(501.70)$ &$(\mathbf{440.60})$ && $(491.2)$ & $(464.6)$
\\
elevator
& $18.91$ &$18.21$ &$13.03$ &$\mathbf{10.87}$ && $17.38$ &$23.99$ &$12.93$ &$\mathbf{10.65}$ && $12.83$ & $10.57$
\\
& $(0.80)$ &$(0.50)$ &$(0.40)$ &$(\mathbf{0.30})$ && $(0.40)$ &$(0.80)$ &$(0.30)$ &$(\mathbf{0.30})$ && $(0.3)$ & $(0.3)$
\\
fried
& $17.15$ &$14.84$ &$6.92$ &$\mathbf{1.28}$ && $12.04$ &$25.03$ &$6.94$ &$\mathbf{1.62}$ && $7.03$ & $1.21$
\\
& $(0.20)$ &$(0.20)$ &$(0.10)$ &$(\mathbf{0.00})$ && $(0.20)$ &$(0.30)$ &$(0.10)$ &$(\mathbf{0.00})$ && $(0.1)$ & $(0.0)$
\\
house-16h
& $24.33$ &$20.21$ &$21.28$ &$\mathbf{9.69}$ && $22.69$ &$27.26$ &$1462200.27$ &$\mathbf{17.52}$ && $22.33$ & $9.88$
\\
& $(2.20)$ &$(1.20)$ &$(1.20)$ &$(\mathbf{1.10})$ && $(1.50)$ &$(1.80)$ &$(963977.00)$ &$(\mathbf{1.50})$ && $(1.9)$ & $(0.8)$
\\
house-8l
& $23.71$ &$20.08$ &$17.95$ &$\mathbf{8.71}$ && $21.16$ &$28.51$ &$197143.11$ &$\mathbf{15.99}$ && $18.04$ & $8.80$
\\
& $(1.50)$ &$(1.10)$ &$(1.40)$ &$(\mathbf{0.40})$ && $(1.20)$ &$(1.30)$ &$(199581.40)$ &$(\mathbf{1.40})$ && $(1.3)$ & $(0.6)$
\\
insurance-charge
& $95.74$ &$91.59$ &$36.46$ &$\mathbf{25.26}$ && $99.42$ &$143.41$ &$68.98$ &$\mathbf{46.36}$ && $40.66$ & $20.72$
\\
& $(8.80)$ &$(8.40)$ &$(4.50)$ &$(\mathbf{3.10})$ && $(10.00)$ &$(13.50)$ &$(7.40)$ &$(\mathbf{7.50})$ && $(6.2)$ & $(3.3)$
\\
kin-8nm
& $566.46$ &$489.96$ &$406.42$ &$\mathbf{192.96}$ && $660.73$ &$689.58$ &$523.07$ &$\mathbf{187.33}$ && $411.87$ & $169.30$
\\
& $(14.40)$ &$(11.40)$ &$(14.10)$ &$(\mathbf{8.00})$ && $(18.70)$ &$(19.60)$ &$(9.30)$ &$(\mathbf{10.00})$ && $(12.4)$ & $(7.4)$
\\
puma-8nh
& $26.68$ &$22.64$ &$19.60$ &$\mathbf{10.77}$ && $26.31$ &$31.56$ &$19.82$ &$\mathbf{10.36}$ && $19.90$ & $10.40$
\\
& $(0.60)$ &$(0.40)$ &$(0.50)$ &$(\mathbf{0.60})$ && $(0.40)$ &$(0.40)$ &$(0.90)$ &$(\mathbf{0.40})$ && $(0.6)$ & $(0.4)$
\\
real-estate
& $155.19$ &$118.97$ &$\mathbf{68.82}$ &$\mathbf{64.47}$ && $130.77$ &$194.92$ &$128.28$ &$\mathbf{61.10}$ && $92.49$ & $54.94$
\\
& $(42.10)$ &$(34.70)$ &$(\mathbf{25.10})$ &$(\mathbf{27.60})$ && $(28.30)$ &$(51.70)$ &$(27.90)$ &$(\mathbf{21.70})$ && $(37.3)$ & $(26.0)$
\\
rmftsa-ladata
& $6.93$ &$5.19$ &$\mathbf{3.77}$ &$\mathbf{4.34}$ && $5.59$ &$8.94$ &$\mathbf{3.89}$ &$5.56$ && $3.99$ & $4.00$
\\
& $(1.00)$ &$(1.00)$ &$(\mathbf{1.00})$ &$(\mathbf{0.90})$ && $(2.10)$ &$(3.50)$ &$(\mathbf{1.60})$ &$(1.20)$ && $(1.1)$ & $(1.1)$
\\
space-ga100
& $313.80$ &$301.18$ &$237.95$ &$\mathbf{127.45}$ && $366.00$ &$384.73$ &$300.72$ &$\mathbf{162.05}$ && $221.42$ & $115.54$
\\
& $(40.60)$ &$(49.40)$ &$(45.60)$ &$(\mathbf{12.50})$ && $(31.30)$ &$(54.70)$ &$(52.80)$ &$(\mathbf{34.20})$ && $(47.8)$ & $(18.7)$
\\
stock
& $26.02$ &$24.76$ &$6.50$ &$\mathbf{1.44}$ && $17.35$ &$43.21$ &$13.19$ &$\mathbf{1.32}$ && $5.86$ & $0.79$
\\
& $(1.90)$ &$(1.30)$ &$(0.50)$ &$(\mathbf{0.30})$ && $(1.10)$ &$(2.40)$ &$(1.00)$ &$(\mathbf{0.20})$ && $(0.7)$ & $(0.1)$
\\
wine-quality
& $0.66$ &$0.62$ &$0.55$ &$\mathbf{0.45}$ && $0.63$ &$0.77$ &$0.55$ &$\mathbf{0.44}$ && $0.55$ & $0.43$
\\
& $(0.00)$ &$(0.00)$ &$(0.00)$ &$(\mathbf{0.00})$ && $(0.00)$ &$(0.00)$ &$(0.00)$ &$(\mathbf{0.00})$ && $(0.0)$ & $(0.0)$
\\
wine-red
& $0.55$ &$0.53$ &$\mathbf{0.40}$ &$\mathbf{0.40}$ && $0.54$ &$0.64$ &$0.44$ &$\mathbf{0.37}$ && $0.41$ & $0.36$
\\
& $(0.00)$ &$(0.00)$ &$(\mathbf{0.00})$ &$(\mathbf{0.00})$ && $(0.10)$ &$(0.00)$ &$(0.00)$ &$(\mathbf{0.00})$ && $(0.0)$ & $(0.0)$
\\
wine-white
& $0.69$ &$0.63$ &$0.56$ &$\mathbf{0.46}$ && $0.61$ &$0.81$ &$0.57$ &$\mathbf{0.45}$ && $0.58$ & $0.44$
\\
& $(0.00)$ &$(0.00)$ &$(0.00)$ &$(\mathbf{0.00})$ && $(0.00)$ &$(0.00)$ &$(0.00)$ &$(\mathbf{0.00})$ && $(0.0)$ & $(0.0)$
\\
\bottomrule
\end{tabular}
} % \resizebox
\end{table}

\end{document}